\documentclass[twoside,11pt]{article}

\usepackage[abbrvbib]{jmlr2e}

\usepackage{amsmath}
\usepackage{amsfonts}
\usepackage{graphicx}
\usepackage{enumerate}
\usepackage{natbib}
\usepackage{url} % not crucial - just used below for the URL 
\usepackage{float}
\usepackage{subfloat}
\usepackage{subfigure}
\usepackage{wasysym}
\usepackage{ulem}
\usepackage{color}

\usepackage{algorithmic}
\usepackage{bm}
\usepackage{natbib}

\usepackage[ruled,noend]{algorithm2e}

\def\d{\partial}
\def\mb{\mathbf}

\def\R{\mathbb{R}}
\def\M{\mathbb{M}}
\def\E{\mathbb{E}} 
\def\V{\mathbb{V}}

\def\X{\mathcal{X}}
\def\S{\mathcal{S}}
\def\G{\mathcal{G}}
\def\U{\mathcal{U}}

\def\IGP{GPUM }

\def\g{\mathbf{g}}
\def\f{\mathbf{f}}
\def\u{\mathbf{u}}
\def\s{\mathbf{s}}
\def\J{\mathbf{J}}

\graphicspath{{fig/}}

\jmlrheading{1}{2000}{1-48}{4/00}{10/00}{meila00a}{ Mu Niu, Zhenwen Dai, Pokman Cheung and Yizhu Wang}

\ShortHeadings{\IGP on Unknown Manifolds}{Niu, Dai, Cheung and Wang}
\firstpageno{1}

\begin{document}

% \title{Intrinsic Gaussian Process Regression on Unknown Manifolds}

% \title{Intrinsic Gaussian Process on Unknown Manifolds with Probabilistic Metric}

\title{Intrinsic Gaussian Process on Unknown Manifolds with Probabilistic Metrics}

\author{\name Mu \ Niu \email mu.niu@glasgow.ac.uk \\
       \addr School of Mathematics and Statistics\\
       University of Glasgow\\
       UK
       \AND
       \name Zhenwen \ Dai \email zhenwend@spotify.com \\
       \addr Spotify, London\\
       UK
       \AND
       \name Pokman \ Cheung \email pokman@alumni.stanford.edu\\
       \addr London, UK
       \AND
       \name Yizhu \ Wang \email 2603214w@student.gla.ac.uk \\
       \addr School of Mathematics and Statistics\\
       University of Glasgow\\
       UK
       }

\editor{}

\maketitle

\begin{abstract}%   <- trailing '%' for backward compatibility of .sty file
This article presents a novel approach to construct Intrinsic Gaussian Processes for regression on unknown manifolds with probabilistic metrics (\IGP) in point clouds. In many real world applications, one often encounters high dimensional data (e.g.‘point cloud data’) centered around some lower dimensional unknown manifolds. The geometry of manifold is in general different from the usual Euclidean geometry. Naively applying traditional smoothing methods such as Euclidean Gaussian Processes (GPs) to manifold-valued data and so ignoring the geometry of the space can potentially lead to highly misleading predictions and inferences. %Intrinsic Gaussian Processes are first introduced by \cite{niu2019} and are only applicable when the geometry of the manifold is known. However, for most real-world problems, data in the point cloud is often high dimensional and its geometry and the parameterisations of the manifolds are unknown. 
%One needs to learn the geometry before utilising it for inference. 
%In this work we extend Intrinsic Gaussian Processes to handle unknown manifolds by estimating the probabilistic parameterisation of the implicit manifolds using probabilistic latent variable models.  
A manifold embedded in a high dimensional Euclidean space can be well described by a probabilistic
mapping function and the corresponding latent space.
We investigate the geometrical structure of the unknown manifolds using the Bayesian Gaussian Processes latent variable models(B-GPLVM) and Riemannian geometry. The distribution of the metric tensor is learned using B-GPLVM. The boundary of the resulting manifold is defined based on the uncertainty quantification of the mapping. We use the the probabilistic metric tensor to simulate Brownian Motion paths on the unknown manifold.  The heat kernel is estimated as the transition density of Brownian Motion and used as the covariance functions of \IGP. The applications of \IGP are illustrated in the simulation studies on the Swiss roll, high dimensional real datasets of WiFi signals and image data examples. Its performance is compared with the Graph Laplacian GP, Graph Mat\'{e}rn GP and Euclidean GP. 
\end{abstract}


\section{Introduction}

Gaussian Processes (GPs) have been widely used as data efficient modelling approaches that produce good uncertainty estimates. They also power many decision making approaches such as Bayesian optimisation, multi-armed bandits and experiment design. Characteristics of the function prior such as differentiability, periodicity and symmetry can be easily controlled by constructing a specific covariance function. Most widely used GP covariance functions are defined over Euclidean space. However, in real world applications, data often lies on a manifold within the original space. Predictions are only valid on the manifold and should only be extrapolated from observations along the manifold. For example, traffic flows can only be measured over networks of roads, surface tension is only measured on the surface of a specific object and the joints of a robot arm can only be moved safely within a manifold of the joint space of all the joints. The traffic connectivity can also easily differ from road network due to road maintenance or traffic congestion. The inference of the manifold based on data brings the accuracy and flexibility. %in the indoor WiFi localization task, the physical location of a mobile device can be determined from wireless signal strengths. The movement of the mobile device is in a constrained complex domain which is the building.??} 
Directly applying a GP with a covariance function defined over Euclidean space would not be ideal as the extrapolation would not respect the geometry of the manifold.

Previously, several GP on manifold methods \citep{niu2019,lin2019,borovitskiy2020} have been proposed under the assumption that the geometry of the manifold is known. \cite{niu2019} defined a heat kernel on a manifold and constructed an intrinsic Gaussian Process on the manifold. The intrinsic GP refers to a GP that employs the intrinsic Riemannian geometry of the manifold, including the boundary features and interior conditions. \cite{lin2019} proposed extrinsic framework for GP modeling on manifolds, which relies on embedding of the manifold into a Euclidean space and then constructing extrinsic kernels for GPs on their images. \cite{dunson2020diffusion} , \cite{borovitskiy2021} and \cite{bolin2022} focused on developing GPs on graphs and metric graphs formed by data observed on the manifold. 
%The heat kernel and Mat\'{e}rn kernel is approximated with finitely-many Eigen pairs of Graph Laplacian. 

In this work, we study GP modelling on manifolds of which the geometry is unknown. This is a more realistic setting for real world applications because measuring the exact geometry of data manifolds is often impossible or overly expensive. In particular, we focus on the scenario where only a sparse set of data points can be collected. As shown in our experiments, Graph Laplacian based methods perform poorly in this scenario due to the poor graph approximation. In contrast, we estimate the probabilistic parameterisation of the unknown manifolds using probabilistic latent variable models and propose a practical and general intrinsic GP on unknown manifolds (GPUM) methodology. This is the major novel contribution of the paper. Specifically, we investigate the geometrical structure of the unknown manifolds using Riemannian geometry. The distributions of the metric tensor and the boundaries of resulting manifolds are estimated using the Bayesian Gaussian process latent variable models (B-GPLVM). Brownian Motion(BM) sample paths on the unknown manifold are simulated using the probabilistic metric and respecting the boundaries. 
The covariance kernel on the unknown manifold is estimated by employing the equivalence relationship between the heat kernel and the transition density of BM on the unknown manifold. We prove this estimator is coordinate independent.
 Our method can incorporate the intrinsic geometry and the uncertainty of the unknown manifold for inference and respect the interior constraints. With numerical experiments on synthetic and real world datasets, we compared our method against Graph Laplacian based methods and GP regressions without manifold estimations. We showed that our method is significantly better than these other methods on all the datasets.

%In our approach, the heat kernel covariance estimation incorporates the geometry and the uncertainty of the probabilistic manifold. This framework allows us to build an intrinsic GP on the unknown manifold (\IGP) by using the heat kernel as the covariance function. The heat kernel is estimated as the transition density of Brownian Motion (BM) on the unknown manifold with probabilistic metric. 

In the following sections, concepts of Riemannian geometry are introduced in Section \ref{sec:concepts}. The metric learning algorithms are explained in Section \ref{sec:metric}. We prove the BM paths on manifolds simulated using different metrics have the same transition density in Section \ref{sec:heatBM}. The heat kernel estimates derived from the analytical metric and different metric learning methods such as Gaussian Processes Latent Variable Model (GPLVM) and B-GPLVM are compared in Section \ref{sec:density}. Applications of \IGP on a synthetic dataset on Swiss roll, high dimensional real datasets of WiFi signals \citep{ferris2007} and COIL images \citep{nene1996} are illustrated and compared to Graph Laplacian GPs \citep{dunson2020diffusion}, Graph Mat\'{e}rn GPs \citep{borovitskiy2021} and Euclidean GPs in Section \ref{sec:simdata}, \ref{sec:wifi} and \ref{sec:real}.
%the BM transition density estimates proposed in this work are coordinate independent
%In this work we extend Intrinsic Gaussian Processes to handle unknown manifolds by estimating the probabilistic parameterisation of the implicit manifolds using probabilistic latent variable models. We investigate the geometrical structure of the implicit manifolds using Riemannian geometry and estimate the distribution of the metric tensor and boundary of the resulting manifolds. 
%Heat kernels on the implicit manifold are used as the covariance functions of the \IGP. The heat kernel is estimated as the transition density of Brownian motion (BM) on the implicit manifold. \IGP can accommodate the interior structure of the implicit manifolds and respect the boundary. 

\section{Concepts of Riemannian Geometry and Theoretical Background}
\label{sec:concepts}

\begin{figure}[t]
   % \centering
    \subfigure[Chart of $\M$(Swiss roll)]{\includegraphics[width=0.7\textwidth,height=0.3\textwidth]{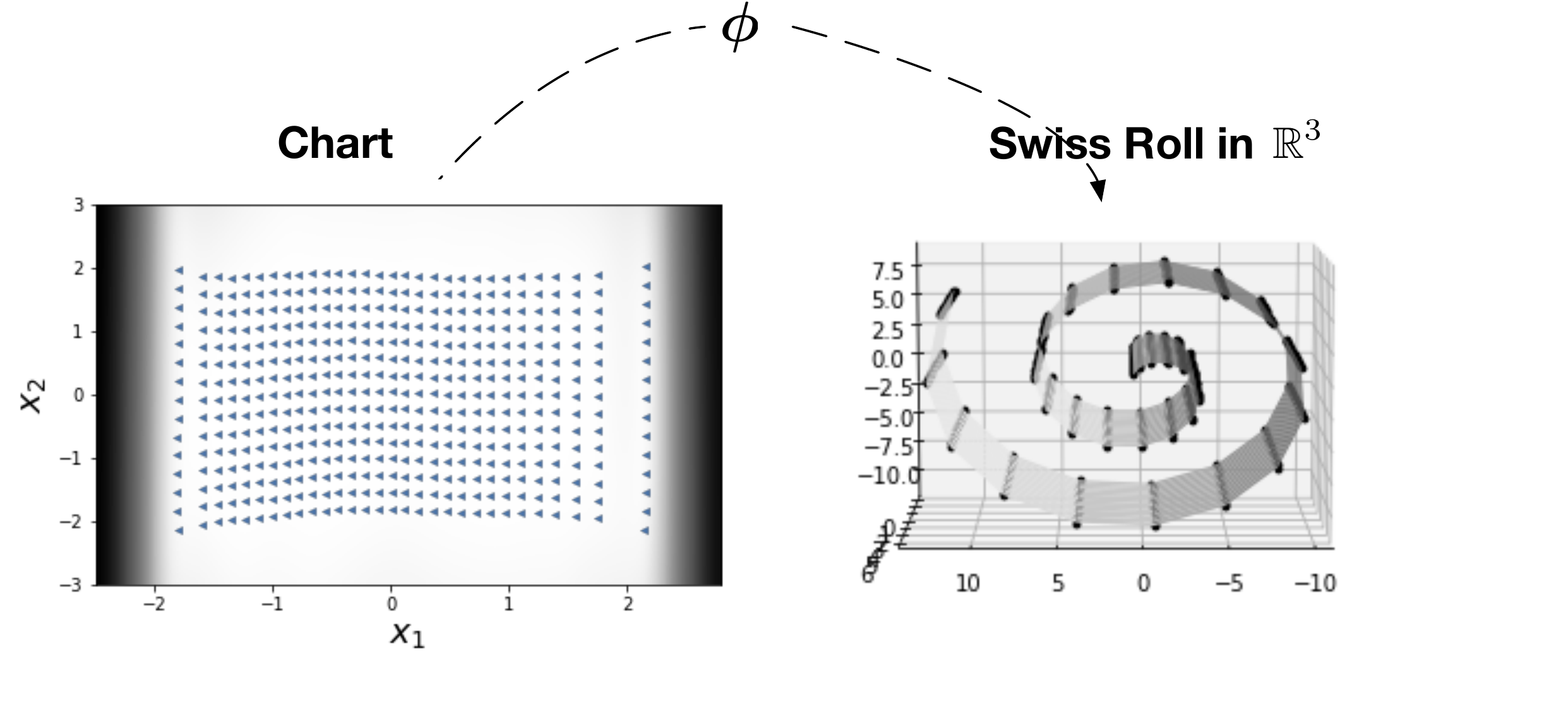} \label{fig:mapping} }
    \subfigure[Regression on $\M$]{\includegraphics[width=0.2\textwidth,height=0.2\textwidth]{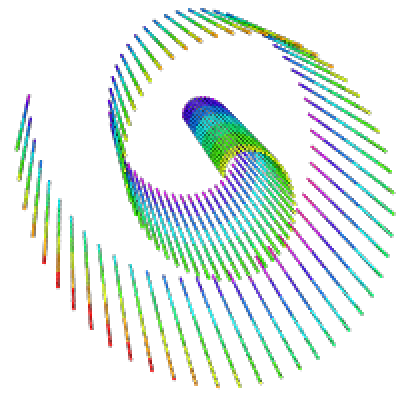} \label{fig:regexample} }
    \caption{ 
    \footnotesize{ 
    { An illustration of a parameterisation of the manifold $\M$(Swiss roll) estimated from a data cloud in (a). $\phi$ maps the chart into $\M \subset \R^3$. The blue triangles in the chart are mapped to the black dots on $\M \subset \R^3$. The Riemannian geometry can help to learn the regression function in (b) (the color indicates function values) %and do not make the inner layer and outer layer of Swiss roll interact
    .}}
    }
\end{figure}

One way of representing a high dimensional dataset is to relate it to a lower dimensional set of latent variables through a set of mapping functions (potentially nonlinear). \cite{tosi2014metric,arvanitidis2019} investigated the geometrical structure of probabilistic generative dimensionality reduction models (or latent variable models) using Riemannian metrics and computed the geodesic distances on the manifolds learned from data. A manifold embedded in a high dimensional Euclidean space can be well described by a probabilistic mapping function and the corresponding latent space. If the dimension of the latent space is the same as the intrinsic dimension of the manifold, the latent space can be interpreted as the chart of the learned manifold. Intuitively, the chart provides a distorted view of the manifold.
An illustration is shown in Fig.\ref{fig:mapping}. The $x^1$ and $x^2$ coordinates in the chart represent the radius and width of the Swiss roll in $\mathbb{R}^3$. The blue triangles in the chart can be mapped to the black points in the embedded space (Swiss roll in $\R^3$)  through $\phi$. In other words, the Swiss roll can be parameterised by the chart. Measurements on the manifold can be computed in the chart locally, and integrated to provide global measures. This gives rise to the definition of a local inner product, known as a Riemannian metric tensor. 

Let $\M$ be a $q$-dimensional complete and compact  Riemannian manifold with the Riemannian metric $\g$, and $\partial \M$ its boundary.  The Riemannian metric $\g$ can be represented as a symmetric, positive definite matrix-valued function, which defines a smoothly varying inner product in the tangent space of $\M$. Let $\mathcal J$ denote the Jacobian of $\phi$. We have %of the $\phi$.
\begin{align} \label{eqn:deter_metric}
\g = \mathcal{J}^T \mathcal{J}, \ \ \  \mathcal{J}_{i,j} = \frac{\partial \phi^i }{ \partial  x^j}
\end{align}
The superscript indicates the $j_{th}$ dimension of the chart and the $i_{th}$ dimension of the observation space.

% ??We first review the background knowledge of heat kernels. 
 Moreover, based on its metric tensor, $\M$ has an associated Laplace-Beltrami operator, which is an intrinsically defined differential operator denoted $\Delta_s$. In local coordinates, the Laplacian-Beltrami operator is%can be written as
\begin{align}
\label{eqn:delta_s}
\Delta_s f = \frac{1}{\sqrt{G}} \frac{\partial}{\partial x^j} \left (    \sqrt{G} \g^{ij} \frac{\partial f}{\partial x^i}  \right ),
\end{align}
where $G$ is the determinant of the metric $\g$, $\g^{ij}$ is the $(i,j)$ element of its inverse and $f$ is a smooth function on $\M$. Take the special case where $\M$ is a Euclidean space $\R^q$, $\g$ becomes an identity matrix. The Laplace-Beltrami operator $\Delta_s$ becomes the Laplace operator $\Delta$ (the sum of second partial derivatives).  
 
Consider the heat equation on $\M$, given by

\begin{align*}
\frac{\d}{\d t}K_{heat}(s_0,s,t)=\frac{1}{2}\Delta_s K_{heat}(s_0,s,t),\qquad  s_0, s\in \M,
\end{align*}
where $s \in \M$, $\Delta_s$ is the Laplacian-Beltrami operator on $\M$, and $t \in \R^{+}$ is the diffusion time. A heat kernel of $\M$ is a smooth function $K(s_0,s,t)$ on $ \M \times \M \times \R^{+}$ that satisfies the heat equation. It can be interpreted as the amount of heat that is transferred from $s_0$ to $s$ in time $t$ via diffusion.
% We first review the background knowledge of heat kernel.

The heat kernel satisfies the initial condition $\lim_{t\rightarrow 0}K_{heat}(s_0,s,t)=\delta(s_0,s)$ with $\delta$ the Dirac delta function. The heat kernel becomes unique when we impose a suitable condition along the boundary $\partial \M$, such as the Neumann boundary condition: $ \partial K / \partial \mathbf{n} = 0$ along $\partial \M$ where $\mathbf{n}$ denotes a normal vector of $\partial \M$. The Neumann boundary condition can be understood as allowing no heat transfer across the boundary. If $\M$ is a Euclidean space $\R^q$,  The heat kernel has a closed form corresponding to a time-varying Gaussian function:
\begin{align*}
K_{heat}(\mb{s}_0,\mb{s},t)
=\frac{1}{(2\pi t)^{q/2}}\,
  \exp\left\{-\frac{||\mb{s}_0-\mb{s}||^2}{ 2t }\right\}, \; \mb{s}\in \mathbb R^q.
\end{align*}
The diffussion time $t$ controls the rate of decay of the covariance. In the following, we will also write $K_{heat}(\mb{s}_0,\mb{s},t)$ as $K_{heat}^t(\mb{s}_0,\mb{s})$.

For arbitrary Riemannian manifold, the construction of the heat kernel associated with the Laplace-Beltrami operator is not a trivial task and belongs to the fields of partial differential equations and differential geometry \citep{chavel1984}. To circumvent solving the heat equation on manifolds, \cite{niu2019} estimates the heat kernel as the BM transition density by simulating BM paths on a known manifold. In the case of point clouds, both the metric tensor $\g$ and the boundary $\partial \M$ are unknown. In this work, we will use probabilistic latent variable models to learn the distribution of the metric $\g$ and define the boundary $\partial \M$ based on the uncertainty quantification of the mapping. The Laplace-Beltrami operator in \eqref{eqn:delta_s} is the infinitesimal generator of BM on manifolds \citep{hsu1988}. The BM on a Riemannian manifold in a local coordinate system is given as a system of stochastic differential equations in the Ito form \citep{hsu1988,hsu2008}:  
\begin{align}
\label{eqnBMhsu}
dx_i(t) = \frac{1}{2}G^{-1/2} \sum^{q}_{j=1}\frac{\partial}{\partial x_j} \left(  \g^{ij}G^{1/2} \right) dt + \left( \g^{-1/2} dB(t)\right)_i,
\end{align}
where $\g$ is the metric tensor of $\M$, $G$ is the determinant of $\g$ and $B(t)$ represents an independent BM in the Euclidean space.

\section{Intrinsic Gaussian Processes on unknown manifolds}
\label{sec:GPmani}

In this work, we focus on the following model, 
\begin{align} \label{eqn:reg}
y_i &= f(s_i) +\epsilon_i, \ \ \  \epsilon_i \sim \mathcal{N}(0, \sigma_{\epsilon}^2) %\\
%s_i &= \phi( x_i ) +e_i \ \ \  e_i \sim \mathcal{N}(0, \sigma_{e}^2) \nonumber
\end{align}
where  $f:\M \rightarrow \R$ is an unknown regression function. $y_i \in \R$ is a response variable. $s_i= (s_{i}^1,\ldots,s_{i}^p) \in \M \subset \R^p$ is an observed predictor on a complete and compact Riemannian manifold $\M$ embedded in $\mathbb{R}^p$.
$\M$ can be parameterised by a $q$ dimensional local coordinate system (chart), and $q<p$. For example, let $\M$ be a unit sphere and we have a point $i$ on $\M$. $s_i \in \R^3$ is a three dimensional vector representing the Cartesian coordinates of $i$. The two dimensional chart of $\M$ is the longitude and latitude coordinate system. The point $i$ can also be represented as a two dimensional vector $x_i$ in the chart. There exists a mapping function $\phi$ from the Cartesian coordinates to the spherical coordinates. However in the case of point clouds where $s_i$ is observed in $\R^p$, the parameterisation $\phi$ of $\M$ is unknown and $x_i$ becomes a latent variable.
%Here $\M$ is implicit, its geometry is not directly observed. 
We propose to infer how the output $y$ varies with the input $s$, including predicting $y$ values at new locations not represented in the training set.

A GP prior can be assigned to $f$ with a covariance function. %$k(s_i,s_j)$. 
The choice of %$k(s_i,s_j)$ 
covariance kernel has a fundamental impact on the results. Most common choices of covariance kernels such as the squared exponential kernel and Mat\'{e}rn kernel depend critically on the Euclidean distance between $s_i$ and $s_j$ and ignore the intrinsic geometry of $\mathbb M$. By contrast, the heat kernel depends only on the Riemmanian metric and the intrinsic geometry of $\M$. It provides a natural generalisation of the RBF kernel on manifolds. The heat kernel represents the diffusion of the heat on a Riemannian manifold. However it is analytically intractable to directly evaluate \citep{hsu1988}. %
\cite{niu2019} proposed a computational framework to estimate the heat kernel on the manifold in which the analytical parameterisation is known. 
However, data represented as point clouds is often high dimensional and concentrated around some unknown lower dimensional structures, \cite{niu2019} is not applicable due to the lack of analytical manifold parameterisation.
In this paper, we address this problem by using the probabilistic generative dimension reduction models to learn the geometry of the implicit manifold and define the boundary. We propose to construct \IGP by using the heat kernel of the implicit manifold as the covariance kernel. With the help of Riemannian geometry, the Brownian motion sample paths can be simulated on $\M$ and the heat kernel can be estimated as the transition density of the BM. With this construction we can learn the regression function on $\M$ as in Fig.\ref{fig:regexample}. 

%{\color{red}By varying the time parameter, one can vary the bumpiness of the realisations of the GPs. }

%We now consider the heat kernel as the covariance function of the \IGP.  

Let $\mathcal{D} =  \{ (s_i, y_i), i = 1, . . . , n \} $ be the data, with $n \geq 1$ the number of labeled observations, $s_i \in \M \subset \R^p$ is the predictor and $y_i \in \R$ is the corresponding response. Suppose we are also given an unlabeled dataset $\mathcal{V} = \{ s_i, i=n+1 . . . ,n+v \} $ where $v \geq 1$. Consider the regression model in \eqref{eqn:reg}, we would like to make inferences about $f$ with the labeled dataset $\mathcal{D}$ and predict $y$ values for the unlabeled dataset $\mathcal{V} $. Under an \IGP prior for the unknown regression function as $f \sim GP(0,K_{heat}^t(.,.) )$, we have
\begin{align*}
p( \mathbf{f} | s_1, s_2, ..., s_n) \sim \mathcal{N} (0, \Sigma_{\f \f}),
\end{align*}
where $\mathbf{f}$ is the discretization of $f$ over $s_1,s_2,\cdot\cdot\cdot,s_n$ so that $f_i = f(s_i)$. $\Sigma_{\f \f} \in \R^{n\times n}$ is the covariance matrix induced from the heat kernel. The $(i,j)$ entry of $\Sigma_{\f\f}$ corresponds to $\Sigma_{\f\f _{i,j}} = \sigma_h^2 K_{heat}^t(s_i,s_j)$. We introduce the rescaling hyperparameter $\sigma_h^2$ to add extra flexibility to the heat kernel.

\section{Related work}
One of the paramount challenges in developing GP models on manifolds is the difficulty in specifying the covariance structure via constructing valid covariance kernels on manifolds. One might hope to achieve this by replacing Euclidean norms in the squared exponential kernel with geodesic distances. However \cite{feragen2015} proved this is not generally a well-defined kernel. \cite{extrinsicGP} proposed extrinsic Gaussian Processes on manifolds by embedding the manifolds onto a higher dimensional Euclidean space. The squared exponential kernel is applied on the images of the manifold after embedding. However, such embeddings are not easy to obtain and only available for certain manifolds when the geometry is known.
\cite{yang2016} proposed a model bypassing the need to estimate the manifold, and can be implemented using standard algorithms for posterior computation in GPs. 
They show that by imposing a GP prior on the regression function with a covariance kernel defined directly on the ambient space (the embedding of the manifold in a high dimensional Euclidean space), the posterior distribution yields a posterior contraction rate depending on the intrinsic dimension of the manifold. They assume that the unknown lower dimensional space where the predictors center around are a class of submanifolds of Euclidean space.  They focus on compact manifolds without boundary. 

\cite{niu2019} proposed to use heat kernels to construct the intrinsic Gaussian Processes on complex constrained domains. Heat kernel can be interpreted as the transition density of Brownian Motion on the manifold \citep{hsu1988}. It is estimated by simulating BM paths on manifolds in which the analytical parameterisation is known in \cite{niu2019}.  
 
Alternatively if the eigen-paires of the Laplacian-Beltrami operator of the manifold are available, \cite{borovitskiy2020} approximated the heat kernel with the sum of finite-many eigen-paires of the Laplacian-Beltrami. Both \cite{borovitskiy2020} and \cite{niu2019} are only applicable when the geometry of the manifold is known and the dimensions of the observation(or embedding) space are low such as $\M \subset \R^2$ or $\M \subset \R^3$ ($p= 2$ or $3$).

%demonstrated the application of In-GPs in modelling the chlorophyll concentration levels in Aral Sea \citep{wood}.  
%It respects the potentially complex boundaries or interior conditions as well as the intrinsic geometry of the spaces.

Most recent research in \cite{dunson2020diffusion} tackled this problem by approximating the heat kernel kernel of a compact Riemannian manifold with finitely-many eigenpairs of the Graph Laplacian using the labeled and unlabeled predictor values. We refer the Gaussian processes constructed by this approximation as the Graph Laplacian Gaussian processes (GL-GPs). Let $\Delta$ be the Laplace-Beltrami operator of a manifold and $ \{ \lambda_i \}_{i=0}^{\infty}$ be the spectrum or eigenvalues of $-\Delta$. Denote $\varphi_i$ the corresponding eigenfunction, for each $i \in N$, we have $\Delta \varphi_i = - \lambda_i \varphi_i$. If the eigen-decomposition of $\Delta$ is known, the corresponding heat kernel of the manifold has the following expression: $K_{heat}(s,s',t) = \sum_{i=0}^{\infty} e^{-\lambda_i t} \varphi_i(s) \varphi_i(s')  $. If the geometry of the manifold and the corresponding $\Delta$ are unknown, $\Delta$ is approximated  by the Graph Laplacian matrix $\bm{L}$ in \cite{dunson2020diffusion}.  $\bm{L}$ is constructed from a point cloud whose adjacency matrix is computed by using a Gaussian function with pair-wised Euclidean distance. The heat kernel is approximated as the summation of finite eigenpairs of $\bm{L}$,  $\sum_{i=0}^{n_G} e^{-\mu_i t} v_i v_i^T $, where $\mu_i$ and $v_i$ are the $i_{th}$ normalised eigenvalue and eigenvector of $\bm{L}$. The implementation of the GL-GP is provided in  Appendix \ref{ax:GL_kern}.  %the kernel estimates from our BM transition density approach in Fig.\ref{fig:swiss-gl-kern} of section \ref{sec:density}

\cite{borovitskiy2021} leveraged the stochastic partial differential equation characterization of Mat\'{e}rn kernel to study their analog for undirected graphs and developed the Graph Mat\'{e}rn Gaussian processes (GM-GPs). The Graph Mat\'{e}rn kernel is constructed with the sum of finitely-many eigenpairs of the normalised Graph Laplacian $\bm{L}$.
$\bm{L}$ is computed from the adjacency matrix of a predefined graph. We implemented the GM-GP by following the instructions in the Github repository in \cite{borovitskiy2021}. The key component of GL-GPs and GM-GPs is the Graph Laplacian. If the true graph connections are not known, the graph constructed based on local distances such as Delaunay triangulation can be error-prone when observations are sparse \citep{hjelle2006}. The graph based methods often result in poor approximation of the manifold when the number of the observations are low. GL-GPs and GM-GPs are applied to the simulation studies and real datasets and compared to GPUM in the later sections.
%?? Since they both use a compact operator(Graph Laplacian) to approximate an unbounded operator(Laplacian-Beltrami), rewrite the eigenvalues are hard to recover and need many points in the point cloud.?? 
%Alternatively if the Laplacian-Beltrami eigen-paires of the manifold are available, \cite{borovitskiy2020} approximated the heat kernel with the sum of finite-many eigen-paires of the Laplacian-Beltrami operator.

There are still some critical gaps in current practices. In particular, the lack of robust methods for carrying out intrinsic statistical inference and  effective models for regressions with manifold-valued data embedded in a point cloud.
In this work we focus on learning the manifold structure using probabilistic dimension reduction methods such as Bayesian GPLVM and constructing the GPUM by using the heat kernel of the learned manifold. Other related methods such as Auto-encoders \citep{kramer1991} and Variational Auto-encoders (VAE)\citep{kingma2019}, can also be considered in this two stage approach. Auto-encoders provide a neural network based framework for learning deep latent variable models.  However, the resulting mapping function is deterministic and the model does not have a built-in quantification of its uncertainty.
There is lack of uncertainty quantification in the learned manifold. Alternatively, VAE
address this concern directly through an explicit likelihood model and a variational approximation of the representation posterior. It learns a generative model by specifying a likelihood of observations conditioned on latent variables and a prior over the latent variables.  Both the likelihood and the variational distributions have parameters predicted by
neural networks that act similarly to the encoder–decoder pair of the classic autoencoder. %The VAE learn a generative model by specifying a likelihood of observations conditioned on latent variables and a prior over the latent variables. 

\section{Learning metrics and boundaries}\label{sec:metric}
Let $\mathcal S = \{ s_i |  i=1,\cdot \cdot \cdot, n+v \}$,  $s_i \in \R^p$, include all predictors in the labeled dataset $\mathcal{D}$ and unlabeled dataset $\mathcal{V}$. We have $\mathcal{S} \in \R^{ (n+v) \times p }$. Suppose we perform probabilistic nonlinear dimensionality reduction by defining a latent variable model that introduces a set of latent (unobserved) variables $\mathcal{X}= \{x_i | i= 1,\cdot \cdot \cdot, n+v \}$, $x_i \in \R^q$, $\mathcal{X} \in \R^{(n+v)\times q}$ and $q < p$. $\mathcal X$ is related to $\mathcal S$ which is observed in a higher dimensional space. A prior distribution is placed on the latent space which induces a distribution over $\mathcal S$ under the assumption of the probabilistic mapping
\begin{align} \label{eqn:mapmodel}
s_{i}^j = \phi^j(x_i) + e_{i}^j,
\end{align}
where $x_i \in \R^q$ is the latent point associated with the $i_{th}$ observation $s_i \in \R^p$ in the original observation space and $i\leq n$. $j$ is the index of the features (dimensions) of $s$ in the observation space and $j\leq p$. $e_{i}^j$ is a Gaussian distributed noise term,   $e_i^{j} \sim \mathcal N(0,\beta^2)$. %$\beta^2$ is the variance. 
An illustration is given in Fig. \ref{fig:mapping} where $q=2$ and $p=3$. If the mapping $\phi$ is linear and the prior $p( \mathcal X)$ is Gaussian, this model is known as probabilistic principal component analysis \citep{tipping1998}. In this work, we do not restrict to this linear assumption and consider some nonlinear dimensionality reduction methods such as GPLVM and B-GPLVM.

When the $\phi$ function in Fig.\ref{fig:mapping} is differentiable, it can be interpreted as the mapping between the latent space and the manifold $\M$. If the dimensions of $\M$ are known, by letting $q$ equal the dimensions of $\M$, the latent space can be interpreted as the chart of $\M$. If the dimensions of $\M$ are unknown, $q$ is estimated by using the so called Automatic Relevance Determination (ARD) in GPLVM and B-GPLVM \citep{zwiessele2017}. ARD allows assigning scaling parameters for each dimension. These scaling parameters can be incorporated to kernels such as the RBF kernel as the inverse of the squared lengthscales.
%When ARD is used, often the input dimensions are sorted based on the relevance assigned by the scaling introduced.

In general %if the topology of a manifold is nontrivial, 
a manifold may need more than one chart to be parameterised. In this work, we focus on examples of single chart. But our method can be extended to learn multiple charts from multiple datasets. %An example of estimating the heat kernel of the cylinder is given in Appendix \ref{apx:cylinder}. 
The Riemannian metric of the given model can be computed as in \eqref{eqn:deter_metric}. In the case of probabilistic LVMs, we place a Gaussian prior over the mapping function $\phi(x)|x$. The conditional probability over the Jacobian also follows a Gaussian distribution, this naturally induces a distribution over the metric tensor $\g$. We denote the distribution of the Jacobian as $\mathbf{J}$ given the set $\mathcal{S}$ and the mapping $\phi$ in \eqref{eqn:mapmodel}. Assuming independent rows of $\mathbf J$ \citep{lawrence2005,titsias2010},
\begin{align}
p(\mathbf J | \mathcal{S}, \Phi) = \prod_{j=1}^p \mathcal{N}( \mu_{\J}^j, \Sigma_{\mathbf J}  ).
\end{align}
This independent row assumption is for the dimensions in the original
observational space. The dimensions in the learned latent space are
not independent. This assumption can be relaxed by using a multi-output GP which is more compuationaly expensive \citep{alvarez2011}. The expressions of the mean $\mu_{\mathbf J}$ and variance $\Sigma_{\mathbf J} $ of the Jacobian are model specific and given in section \ref{gplvm} and \ref{bgplvm}. The resulting metric $\g$ follows a non-central Wishart distribution \citep{anderson1946}
\begin{align} \label{eqn:metric}
\g \sim \mathcal{W}_q \left(  p, \Sigma_{\mathbf{J}}, \E(\mathbf{J}^T)  \E(\mathbf{J})  \right).
\end{align}
From this distribution, the expected metric tensor can be computed as
\begin{align} \label{eqn:exmetric}
\mathcal{G} = \E( \g )  = \E( \mathbf{J}^T) \E( \mathbf{J} ) + p \Sigma_{\mathbf{J}}. % \E(\mathbf{J}^T \mathbf{J})
\end{align}
We denote the expectation by $\mathcal{G}$. Note that the variance term $\Sigma_{\mathbf{J}}$ is included in $\mathcal{G}$. It implies that the metric tensor expands as the uncertainty over the mapping increases.
Hence the BM simulation steps in the SDE in (\ref{eqnBM}) will travel `slower' in the region of the latent space where the uncertainty is high. 

We also need the gradient of the expected metric to simulate BM as in section \ref{trandens}.
\begin{align}
\frac{\partial \G}{\partial x^l} = \frac{ \partial \E[ \g]}{\partial x^l} = \frac{\partial \E[\J^T] } { \partial x^l }  \E[\J] + \E[\J^T] \frac{ \partial  \E[\J]  }{\partial x^l}  + p \frac{ \partial \Sigma_\J }{ \partial x^l }
\end{align} 
The estimates of the mapping  $\phi(x)$ become highly unreliable when $x$ is far from the data points. The corresponding metric tensor estimates and BM simulations can also be ill defined and violate the manifold geometry. To avoid these poorly estimated region due to lack of data, the boundary of the learned manifold can be defined by $\V ar( \phi(x) | x)$, the variance of the mapping at $x$. $\V ar( \phi(x) | x)$ is also a smooth function in B-GPLVM. In most cases, the $\partial \M$ defined here is also a $q-1$ smooth manifold. Any $x$ outside of the boundary has $\V ar( \phi(x) | x ) > \alpha$.
\begin{align}\label{eqn:bound}
\partial \M = \{  x \in \R^q \  \vert \ \  \V ar( \phi(x) | x) = \alpha   \}.
\end{align}
where $\alpha$ is defined as the maximum variance of the mapping at the shifted latent points, $\max \left( \V ar\left( \phi(h) | h \right) \right)$. The shifted latent points are randomly sampled from the set $\mathbb{H}=\{ h |  \| h -x_i\| = \delta_{\X}, x_i\in \mathcal{X}  \}$.  $\delta_{\X}$ can be chosen according to the maximum distance of two neighbouring latent points. In practice, we create $h$ by moving all data points with $\delta_{\mathcal{X}}$ in random directions and let $\alpha$ equal to the maximum variance of the mapping at these relocated points. The samples are chosen to make sure that $\alpha > \max \left( \V ar\left( \phi(x_i) | x_i \right) \right)$, for all $x_i\in \mathcal{X}$. %To make sure $\alpha$ is bigger than the variance of mapping at any data points, we can keep resampling the `shifted latent points' until $\alpha > \max \left( Var\left( \phi(x_i) | x_i \right) \right)$, for all $x_i\in \mathcal{X}$.

\subsection{ GPLVM metric}\label{gplvm}
Gaussian Process Latent Variables Model \citep[GPLVM; ][]{lawrence2005} is a nonlinear probabilistic generative model. In this section, we will derive the distribution of the metric from GPLVM.
A sample from GPLVM defines a generative mapping from $x\in \R^q$ in the latent space  to $s \in \M \subset \R^p$ in the observation space.
GPs define a prior over the mapping $\phi$ in \eqref{eqn:mapmodel}. A zero mean prior is used as a default choice. If the domain
knowledge of where the prior should be centred is available, such as the embedding of $\R^q$ into $\R^p$, it can also be encoded into
the mean function. The individual dimensions of the $p$ dimensional observation space are modeled independently in the GP prior sharing the same hyperparameters. Given the construction outlined above, the probability of the observed data $\S =\{ s_i^j| i\in \{1,\cdots,n+v \},j\in \{1,\cdots,p\}  \}$ conditioned on all latent variables $\X = \{ x_i^j| i\in \{1,\cdots,n+v \},j\in \{1,\cdots,q\}   \}$ is written as follows:
\begin{align}\label{eqn:gplvmJoint}
p(\S, \Phi |\X , \beta) =   p(\S | \Phi ,\beta ) p(\Phi | \X)   = \prod_{j=1}^p p(\s_{:}^j | \bm{\phi}_{:}^j, \beta ) p( \bm{\phi}_{:}^j | \X),
\end{align} 
 where $\Phi = \{ \bm{\phi}_i^j | i\in \{ 1,\cdot\cdot\cdot, n+v \} , j\in \{ 1,\cdot\cdot\cdot, p \} \} $, $\Phi \in \R^{(n+v)\times p}$ and $ \bm{\phi}_i^j = \phi(x_i)^j$.  $\s_{:}^j$ represents the $j_{th}$ dimension of all points in $\mathcal S$ and $\mathcal S \in \R^{ (n+v)\times p}$. The likelihood $p( \mathcal S | \mathcal{X} )$ is computed by marginalising out $\Phi$ and optimising the latent variables $\X$
\begin{align}
p(\S | \X ) = \prod_{j=1}^p \mathcal N( \s_{:}^j, K_{\X \X} + \beta^2 I ).
\end{align}
$K_{\X \X} $ is the $(n+v)\times (n+v)$ covariance matrix defined by the squared exponential kernel. 
%Since $x \in \X$ is a latent variable, we can assign it a prior density given by the standard normal density. 
\cite{lawrence2005} estimated all the latent variables $\mathcal X$ and the kernel hyper-parameters of GPLVM with the maximum likelihood estimate. If the covariance kernel is differentiable, the derivative of a GP is again a GP \citep{Rasmussen2004}. This property allows us to compute the derivative of GP. The Jacobian $\J$ of the GPLVM mapping can be computed as the partial derivative $\frac{\partial \phi_*}{ \partial x^l}$ with respect to the $l_{th}$ dimension for any point $x_*$  in the latent space, 
\begin{align} \label{eqn:Jacobian}
\J^{T} = \frac{\partial \phi_*}{ \partial x} = \begin{bmatrix}
 \frac{ \partial \phi(x_*)^1 }{\partial x^1}   &   \cdot \cdot \cdot  &    \frac{ \partial \phi(x_*)^j }{\partial x^1}   &   \cdot \cdot \cdot  &    \frac{ \partial \phi(x_*)^p }{\partial x^1}   \\
  \cdot \cdot \cdot  &  \cdot \cdot \cdot  &  \cdot \cdot \cdot  &  \cdot \cdot \cdot  &  \cdot \cdot \cdot \\
  \frac{ \partial \phi(x_*)^1 }{\partial x^q}   &   \cdot \cdot \cdot  &    \frac{ \partial \phi(x_*)^j }{\partial x^q}   &   \cdot \cdot \cdot  &    \frac{ \partial \phi(x_*)^p }{\partial x^q}
\end{bmatrix},  
\end{align}
where $\frac{\partial \phi_*}{ \partial x}$ is a $q \times p$ matrix. Considering the independence across the dimensions of the observation space, the joint distribution of the $j_{th}$ dimension of the mapping $\phi$ and the $j_{th}$ column of the Jacobian can be written as 
\begin{align}
\begin{bmatrix}
 \phi( \X)^j\\
 \frac{ \partial \phi(x_*)^j }{\partial x} % \frac{ \partial y(x,t) }{\partial x}
\end{bmatrix}, \
\sim \mathcal N \left( 0, \begin{bmatrix}
K_{\X,\X} & \partial K_{\X,*} \\
 \partial K_{\X,*}^T & \partial^2 K_{*,*}
\end{bmatrix}  \right ).
\end{align}
The expressions of $K_{\X,\X}$ ,  $\partial K_{\X,*}$ , and  $\partial^2 K_{*,*} $ are given in Appendix \ref{ax:GPLVM}. GPLVM provides an explicit mapping from the latent space to the observation space. This mapping defines the support of the observed data $\mathcal{S}$ as a $q$-dimensional manifold embedded in $\R^p$. The distribution of the Jacobian of GPLVM is the product of $p$ independent Gaussian distributions (one for each dimension of the observation space). For a point $x_*$ in the latent space, the  distribution of the Jacobian takes the form %\ref{ax:GPLVM}
\begin{align}
p(  \J | \mathcal{X}, \mathcal{S}) &= \prod^p_{j=1} \mathcal N( \mu_{\J}^j , \Sigma_{\J} ) \\
   &=  \prod^p_{j=1} \mathcal N( \partial K^T_{\X,*} K^{-1}_{\X,\X} \s_{:}^j  \ , \partial^2K_{*,*} - \partial K^T_{\X,*} K_{\X,\X}^{-1} \partial K_{\X,*}  ). \nonumber
\end{align} 
From this distribution, the expected metric tensor can be computed as in \eqref{eqn:exmetric}. The boundary $\partial \M$ is defined by $\V ar( \phi(x_*) | x_*)$, the variance of mapping in \eqref{eqn:bound}. For any $x_*$ in the latent space
\begin{align}\label{eqn:varb}
\V ar(\phi(x_*) | x_*) = K_{*,*} - K_{*,\X}K_{\X,\X}^{-1} K_{\X,*}.
\end{align}

\subsection{Bayesian GPLVM metric }\label{bgplvm}
%The GPLVM based models rely on maximum likelihood training procedures for optimising the latent inputs and the hyper-parameters \citep{lawrence2005}. 
The maximum likelihood estimation of the latent inputs  $\X$ in GPLVM often leads to overfitting due to its high dimensionality \citep{damianou2016}. 
This overfitting can be avoided by applying a Bayesian treatment to the latent inputs. By introducing a prior distribution to the latent inputs, the marginal likelihood takes the form
\begin{align*}
p(\mathcal S)=\int p(\mathcal{S} \mid \X) p(\X) d \X.
\end{align*} 
The  integral is intractable as the inputs $\X$ to the latent variable model go through a non-linear calculation in the inverse of the covariance matrix. Bayesian Gaussian Process Latent Variable Model \citep[B-GPLVM;][]{titsias2010} introduces a variational inference framework for training the latent variable model. It variationally integrates out the input variables and computes a lower bound on the exact marginal likelihood of the nonlinear latent variable model. The maximization of the variational lower bound provides a Bayesian training procedure that is robust to overfitting. 

The key to the tractable variational Bayes approach is the application of variational inference to an augmented GP formulation, known as sparse GP, where the GP prior on $\phi$ is augmented to include auxiliary variables. 
More specifically, we expand the conditional probabilistic model in \eqref{eqn:gplvmJoint} by including $m$ extra samples (inducing points) to the GP latent mapping %$\phi(x)$, 
e.g. $u_i = \phi(x_{ui}) \in \R^p$ is such a sample \citep{lawrence2007}.
These inducing points are denoted by $\mathcal{U} = \{u_i^j |  i \in \{ 1,\cdot\cdot\cdot,m \},  j \in \{  1,\cdot\cdot\cdot,p \}    \}$, $\mathcal{U} \in \R^{m\times p}$ and constitute latent function evaluations at a set of pseudo-inputs $\X_u = \{ x_{ui}^j | i \in \{1,\ldots,m \} , j \in \{ 1,\ldots,q\} \} $, $\X_u \in \R^{m \times q}$. 
The inducing inputs $\X_u$ are variational parameters. 
$\Phi$ is defined as in \eqref{eqn:gplvmJoint}. The augmented joint probability and the marginal likelihood can be written as \eqref{eqn:full-marginal}
\begin{align}\label{eqn:full-marginal}
p(\S , \Phi , \mathcal{U} ,  \X | \beta) &=   p(\S | \Phi ) p(\Phi | \mathcal{U} ,\X) p(\mathcal {U}) p(\X)  = \prod_{j=1}^p p(\s_{:}^j | \bm{\phi}_{:}^j) p( \bm{\phi}_{:}^j | \u_{:}^j, \X) p(\u_{:}^j) p(\X), \\
p(\S) &= \int  \int  \int  p(\S, \Phi, \mathcal{U}, \X) \ d \mathcal{U}     \  d\Phi \ d \X. 
\end{align} 
%= \int \int  \int  \left( \prod_j^p p(y_{:,j} \mid \bm{\phi}_{j} ) p(\bm{\phi}_j \mid \mbu_j,X ) \ d\bm{\phi}_j \  p(\mbu_j) d\mbu_j \right) \  p(X) dX \\
%\int \int  \int  p(S \mid \phi) p(\phi \mid u,X ) \ df \  p(X) dX \  p(u) du \nonumber  \\
%&= \int \int  \int  \left( \prod_j^p p(s_{:}^j \mid \bm{\phi}^j ) p( \phi^j \mid \mbu_j, \X ) \ d\bm{\phi}_j \  p(\mbu_j) d\mbu_j \right) \  p(\X) d\X 
With variational inference, $\log \left(p(\S )\right)$ can be lower bounded by applying Jensen’s inequality \citep{damianou2016}. 
The resulting lower bound can be computed analytically. The full technical details of the lower bound are given in \eqref{apxeqn:vlowbou} in Appendix \ref{ax:BGPLVM}. The Jacobian has the same shape as in \eqref{eqn:Jacobian}. For a point $x_*$ in the latent space, the distribution of the Jacobian for B-GPLVM takes the form: %In contrast to the GPLVM, B-GPLVM variationally integrates the latent variables $\X$ out.  \ref{ax:BGPLVM}
\begin{align}
p(  \J | \X,\mathcal{U},\S) &= \prod^p_{j=1} \mathcal N( \  \mu_{\J}^j , \Sigma_{\J} ) \\
   &=  \prod^p_{j=1} \mathcal N( \  \partial K^T_{\X_u,*} K^{-1}_{\X_u,\X_u} \mu_{qu}^j , \  \partial^2K_{*,*} - \partial K^T_{\X_u,*} \Lambda \partial K_{\X_u,*}  )  \nonumber   \\
 \Lambda &= K_{\X_u\X_u}^{-1} - K_{\X_u\X_u}^{-1} \Sigma_{qu} K_{\X_u\X_u}^{-1} \nonumber  \\
\V ar(\phi(x_*) | x_*) &= K_{*,*} - K_{*,\X_u} \Lambda K_{\X_u,*}\label{eqn:varb}
\end{align} 
where $\mu_{qu}^j$ and $\Sigma_{qu}$ are the mean and variance of the variational distribution of the inducing points $\mathcal{U}$. The expressions of $\mu_{qu}$ and $\Sigma_{qu} $ are given in \eqref{apxeqn:vquqsig} in Appendix \ref{ax:BGPLVM}. The expectation of the metric tensor can be computed as in \eqref{eqn:exmetric}. The boundary $\partial \M$ of the implicit manifold can be defined by computing the variance of the mapping as in \eqref{eqn:varb}. %Appendix \ref{ax:BGPLVM}

\section{Estimate heat kernel as BM transition density} \label{sec:heatBM}

 In this section, we will estimate the BM transition density by simulating BM sample paths on the implicit manifold $\M$. %Let $\mathbf{S}_t$ be the BM with transition density $K_{heat}^t$. This transition density also satisfies the heat equation.

\subsection{ Simulating Brownian motion on implicit manifolds }
\label{trandens}
\begin{figure}[t]
    \centering
\includegraphics[width=0.7\textwidth,height=0.3\textwidth]{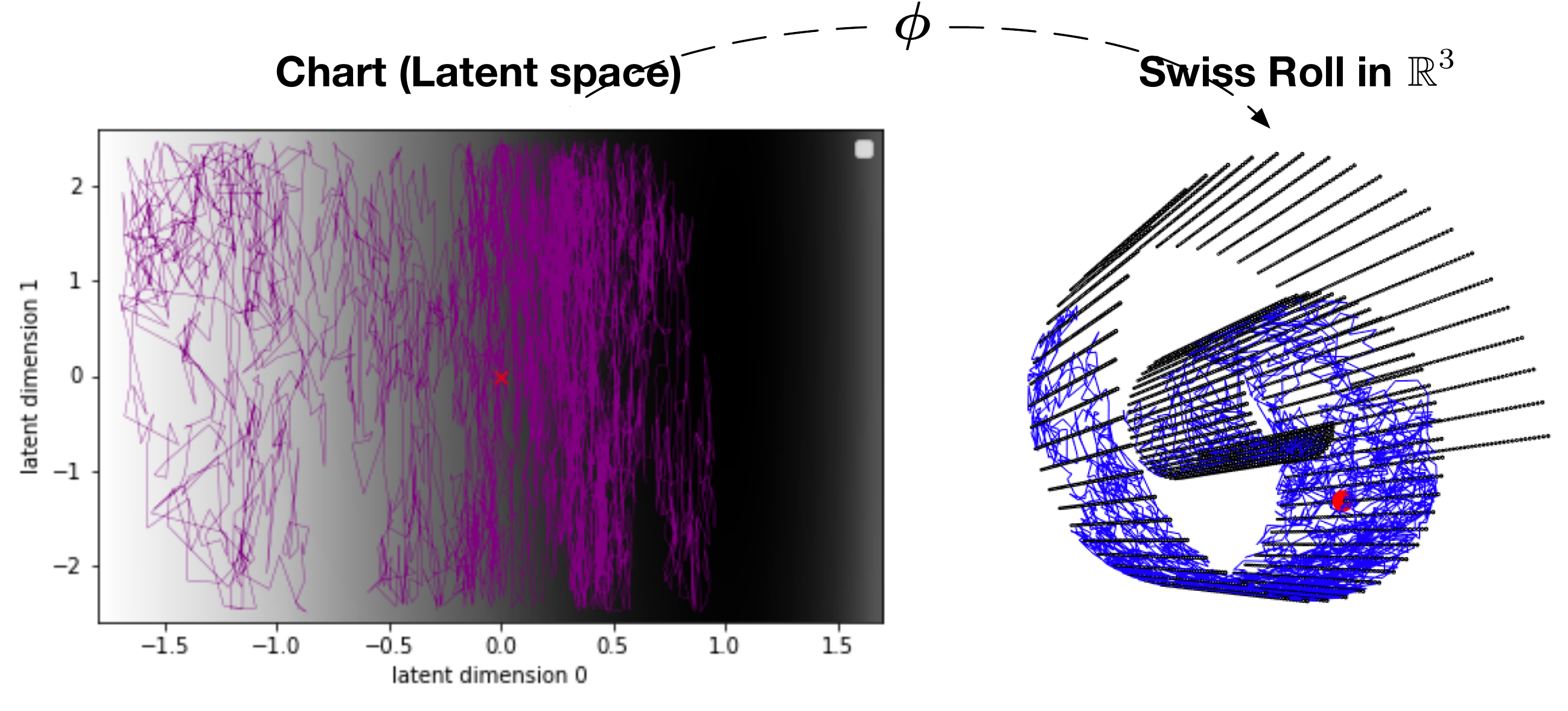}
    \caption{ \label{fig:BM}
    \footnotesize{ 
    {A BM sample path (blue line, right panel) on $\M$ (Swiss roll in $\R^3$) and its equivalent stochastic process (purple line, left panel) in the chart in $\mathbb{R}^2$. $\phi: \mathbb{R}^2 \rightarrow \M \subset \R^3$ is a parametrization of $\M$. The red dot is the starting location of the BM trajectory. The horizontal axis of the latent space represents the radius of Swiss roll. The vertical axis is for the width. The gray color in the latent space indicates the magnification factor. When the latent space is mapped to $\M \subset \R^3$, the darker region will be stretched more.}}  
    }
\end{figure}

From section \ref{sec:metric}, we learned the probabilistic parameterisation $\phi$ of $\M$, the distribution of the associated metric tensor $\g$ and the boundary $\partial \M$. In order to estimate the BM transition density on $\M$, we first need to simulate BM trajectories. Simulating the sample paths of BM on $\M \subset \R^p$ is equivalent to simulating the stochastic processes in the latent space (or chart) in $\R^q$, $q<p$. An illustrated example is shown in Fig. \ref{fig:BM}.
%Let $\mathbf{s}_0$ be the starting point of the Brownian Motion $\mathbf{S}_t$ on manifold $\M$, 

BM on a Riemannian manifold in a local coordinate system is given as a system of stochastic differential equations (SDE) in Ito form \citep{hsu1988}. We use the expected metric $\G = \E[\g]$ to construct the SDEs 
\begin{align}
\label{eqnBM}
dx^i(t) = \frac{1}{2}G^{-1/2} \sum^{q}_{j=1}\frac{\partial}{\partial x^j} \left(  \G^{ij}G^{1/2} \right) dt + \left( \G^{-1/2} dB(t)\right)_i
\end{align}
where $x^i$ represents the $i_{th}$ dimension of the latent space (chart). $\G$ is defined in eqn (\ref{eqn:exmetric}), %the expected metric tensor of $\M$, 
$G$ is the determinant of $\G$ and $B(t)$ represents an independent BM in Euclidean space. The discrete form of \eqref{eqnBM} is derived in \eqref{disBM}.
\begin{align}
\label{disBM}
x^i(t) &= x^i(t-1) + \frac{1}{2} \sum^{q}_{j=1} \left(  -\G^{-1} \frac{\partial \G}{\partial x^j} \G^{-1} \right)_{ij} \Delta t + \frac{1}{4} \sum_{j=1}^q(\G^{-1})_{ij} tr\left(\G^{-1}\frac{\partial \G}{\partial x^j}\right) \Delta t + \left( \G^{-1/2} dB(t)\right)_i \nonumber \\
         &= \mu( x^i(t-1),\Delta t)_i + \left( \sqrt{\Delta t} \G^{-1/2} z^q\right)_i
\end{align}
where $\Delta t$ is the diffusion time of the BM simulation step and $z^q$ is a $q$-dimensional standard normal random variable. The discrete form of the SDE also defines the proposal mechanism of the BM
\begin{align}
\label{eqn:disProp}
q\left(x(t)|x(t-1)\right) = \mathbb N \left( x(t) |  \mu \left( x\left(t-1\right),\Delta t \right)  , \Delta t \G^{-1} \right ).
\end{align}

\begin{theorem} \label{theorem:sde}
The stochastic process defined in \eqref{eqnBM} is coordinate independent. With a given $\delta t$, simulations in any choice of local coordinates (or metric) as above are equivalent to the same step in $\M$.
\end{theorem}

The proof of Theorem \ref{theorem:sde} is given in Appendix \ref{ax:independent}. This implies that the BM sample paths simulated from different choices of metric $\G$ should have the same properties. The BM steps are sampled from the proposal distribution in \eqref{eqn:disProp}, which is defined by the metric tensor. The boundary $\partial \M$ of the manifold is also quantified by the uncertainty of the mapping. We apply the Neumann boundary condition as in Section \ref{sec:GPmani}. As a result the simulated sample paths stay within the boundary. An example of BM trajectory on Swiss roll is given in Fig. \ref{fig:BM}. The latent space and the associated metrics are learned from B-GPLVM. The gray color in the latent space indicates the square root of the determinant of the metric. It is also known as the magnification factor \citep{bishop1997,zwiessele2017,tosi2014}. %\ref{ax:independent}
\begin{align*}
\mathcal{MF} = \sqrt{det(\G) }
\end{align*}
The geometric interpretation of the magnification factor is how much a small piece of the latent space in $\R^q$ will be stretched or compressed when it is mapped to $\M \subset \R^p$. For example in Fig.\ref{fig:BM} left panel, the horizontal axis of the latent space can be interpreted as the scaled radius of the Swiss roll, the vertical axis as the width of the Swiss roll. When the radius is bigger, the corresponding area in the latent space is darker and the magnification factor is larger. When the latent space is mapped back to the manifold, the darker region will be stretched more. The purple trajectory of the stochastic process in the latent space (Fig. \ref{fig:BM} left panel) is denser in the darker area and more spread out in the bright area. The stochastic process travels with smaller steps when the magnification factor is large and vice versa. 
Note that, due to the BM being coordinate independent as in Theorem \ref{theorem:sde}, the BM trajectory is evenly spread out in the manifold of the Swiss roll in $\R^3$ (see Fig. \ref{fig:BM} right panel).
%travels with bigger step when the magnification factor is small. 

\subsection{Estimate the transition density of BM} \label{sec:density}
\begin{figure}[t]
 \centering
\includegraphics[width=0.45\textwidth,height=0.4\textwidth]{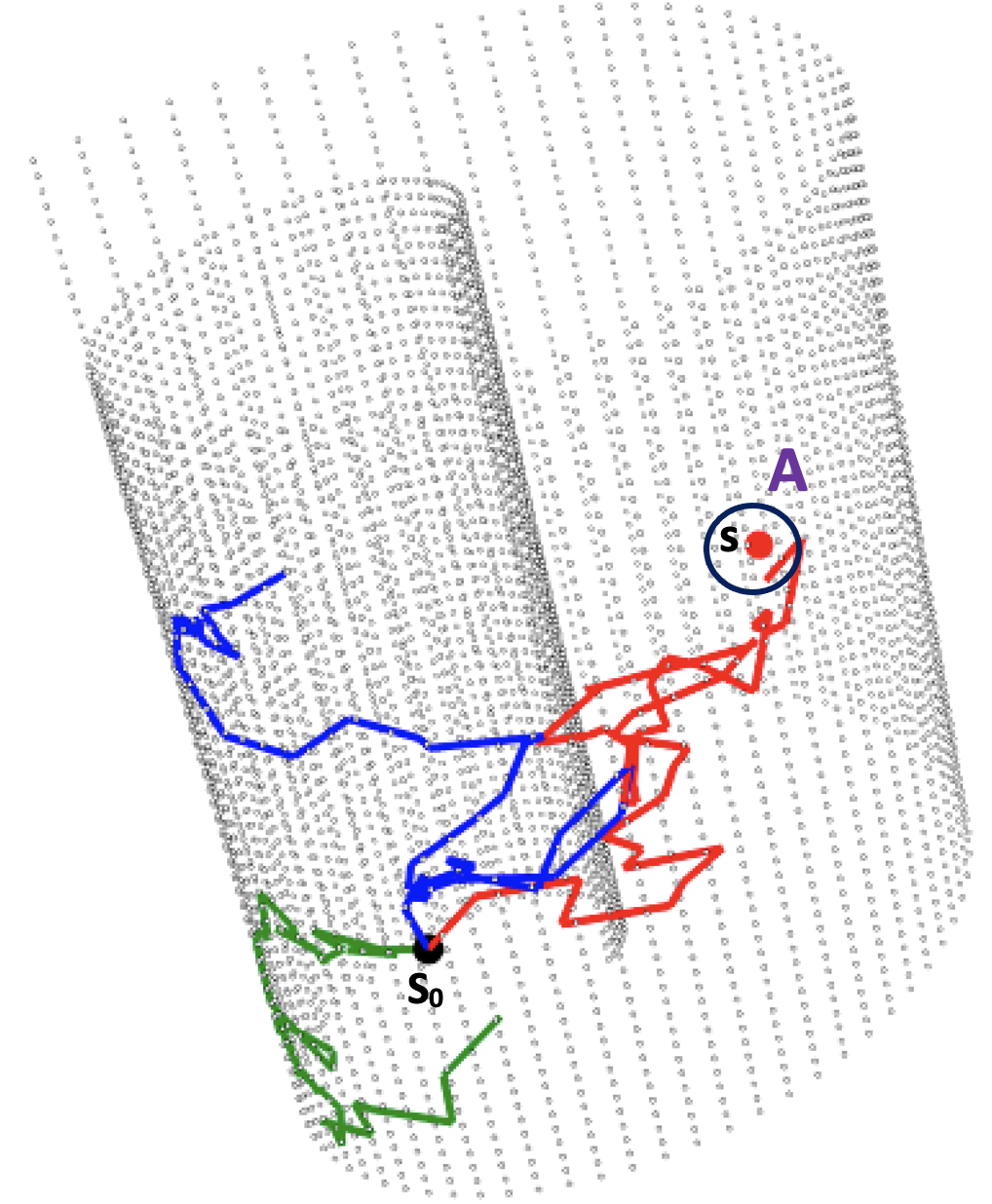}
    \caption{ \label{fig:BM-density}
       {\footnotesize 
    Three BM paths (red,blue,green) are simulated on the Swiss Roll, with the starting point at $s_0$ (black ball). $s$ (red ball) is the target point. $\bm{A}$ is the neighbourhood of $S$. Only the red path reach $\bm{A}$ at time $t$. The transition probability $p\{S(t)\in \bm{A} | S(0)=s_0 \}$ is 1/3.
     } }
\end{figure}

Considering the BM $\{ \mathbf{S}(t) | t>0 \}$ on $\M \subset \R^p$. The BM starts from $\mathbf{S}(0)= s_0$ at time $0$. We simulate $N_{BM}$ sample paths. Given a point $s \in \M$, we define a small neighbourhood of $s$ as $\mathbf{A}_s \in \M$. For any $t > 0$, the probability of $\mathbf{S}(t)$ reaching $\mathbf{A}_s$ at time t, $p( \mathbf{S}(t)\in \mathbf{A}_s | \mathbf{S}(0)=s_0 )$, can be approximated by
\begin{align}
\label{eqn:BMdensity}
p( \mathbf{S}(t)\in \mathbf{A}_s\,| \ \mathbf{S}(0)=s_0)  \approx  \frac{N_{A_s}}{N_{BM}}
\end{align}
where $N_{A_s}$ is the number of sample paths reaching $\mathbf{A}_s$ at time $t$. An illustrative diagram is shown in Fig. \ref{fig:BM-density}. This BM transition probability is defined as the integral of the BM transition density over $\mathbf{A}_s$ which is also the heat kernel $K_{heat}^t$.
%\begin{align}
%\label{BMprob}
%p( \mathbf{S}(t) \in \mathbf{A}_s \ | \ \mathbf{S}(0)=s_0) = \int_{A_s} K_{heat}^t(s_0,s) ds
%\end{align}
%The integral is defined with respect to the volume form of $\M$. 
Since we do not have the analytical expression of the transition probability,  the transition density cannot be derived by taking the derivative of $p( \mathbf{S}(t)\in \mathbf{A}_s | \mathbf{S}(0)=s_0 )$. Instead, $K_{heat}^t$ can be numerically approximated as $ \hat{K}_{heat}^t $
\begin{align}
\label{k:heat}
K_{heat}^t(s_0,s) \approx  \frac{p( \mathbf{S}(t) \in \mathbf{A}_s \ | \ \mathbf{S}(0)=s_0 )}{V(\mathbf{A}_s)} \approx \frac{1}{V(\mathbf{A}_s)}\cdot \frac{N_{A_s}}{N_{BM}} = \hat{K}_{heat}^t 
\end{align}
where  $V(\mathbf{A}_s)$ is the Riemannian volume of $\mathbf{A}_s$. $V(\mathbf{A}_s)$ is parameterised by its radius $\omega$. \cite{niu2019} has provided a indication of the optimal order of magnitude of $\omega$ by minimizing the error of the kernel estimator. When $V(A_s)$ is large, the error of estimating the transition probability becomes smaller. But the error of approximating the transition density become bigger. The former is called Monte Carlo error and the later is called numerical error in Niu et. al (2019). The optimal order of magnitude of $\omega$ can be derived by balancing these two errors.

The Neumann boundary condition corresponds to BM reflecting at the boundary. This can be approximated by pausing time and resampling the next step until it falls into the interior of $\M$. This estimator is asymptotically unbiased and consistent \citep{niu2019}.  Note that $t$ is the BM diffusion time.  If $t$ is large, the BM paths have a higher probability to reach the neighbourhood of the target point leading to higher covariance and vice versa. The transition density can be estimated using Algorithm \ref{alg:bm}.

\begin{corollary} \label{corollary:metric}
The transition density estimate in \eqref{k:heat} is coordinate independent. %independent of the choice of coordinate system.
\end{corollary}
By Theorem \ref{theorem:sde}, it is straightforward to have Corollary \ref{corollary:metric}. The transition density estimate in \eqref{k:heat} of the stochastic process defined in \eqref{eqnBM} is also coordinate independent. 
%Based on corollary \ref{corollary:metric}, we can compare the heat kernel estimates from different parameterisations of the same manifold. 
Based on Corollary \ref{corollary:metric},  we can evaluate the heat kernel of the implicit manifold estimated by different LVMs, independent of the specific parameterisations of their latent spaces. 
If the estimated manifold is close to the true manifold, we expect the resulting BM transition density estimates to be similar to the ones estimated using the analytical parameterisation.

\begin{figure}[h]
 \centering
%\subfigure[ ]{ \label{fig:metriccompare} \includegraphics[width=0.44\textwidth,height=0.3\textwidth]{metricCompare.pdf} }
%\subfigure[ ]{ \label{fig:swiss-gl-kern}  \includegraphics[width=0.44\textwidth,height=0.3\textwidth]{GLkernel.pdf} }
\includegraphics[width=0.6\textwidth,height=0.5\textwidth]{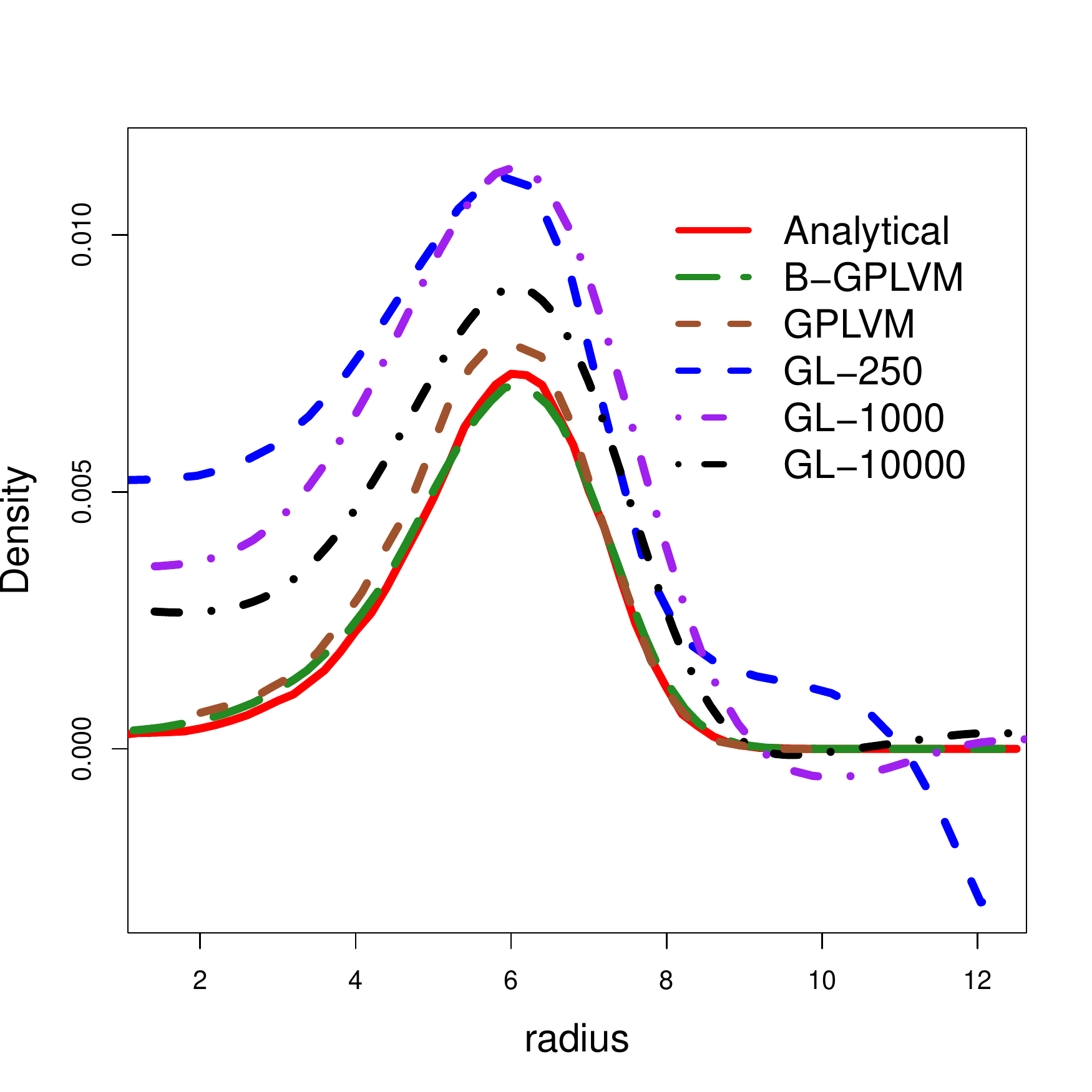}
    \caption{ \label{fig:swiss-gl-kern}
       {\footnotesize 
Comparison of heat kernel estimates using the analytical metric, B-GPLVM metric, GPLVM metric and Graph Laplacian.  The red solid line represents heat kernel estimates using the analytical metric. The green dashed line represents estimates using B-GPLVM metric. The brown dash dotted line represents estimates using GPLVM metric. The heat kernel estimates from the Graph Laplacian approach are plotted as the blue dashed line. We increase the number of the grid points on the Swiss roll from 250 to 1000 and 10000. The GL estimates are plotted as the purple dotted line and black dashed line. 
     } }
\end{figure}

%Estimates of the heat kernel (or BM transition density) with different parametrisations are shown in Fig. \ref{fig:metriccompare}. We repeat this procedure with the analytical metric multiple times. The mean density estimates plus and minus two times the standard deviation are plotted as the black dotted lines.     The green dashed line in Fig.\ref{fig:metriccompare} represents the heat kernel estimates using the metric learned by Bayesian GPLVM. The blue dotted line is for GPLVM. It is clear that the Bayesian GPLVM results are very close to the red solid line. The green dashed line is located between the two black dot lines.     The root mean squared differences between the density estimates using the analytical metric and the density estimates using estimated metrics are also shown in Table \ref{tb:metriccompare}.
%The transition densities are underestimated in the GPLVM results compared to the analytical ones. 
Here we take the Swiss roll as an example. Assuming the geometry of the Swiss roll is known, we can follow \cite{niu2019}'s approach and evaluate the heat kernel $K_{heat}^t(s_0,s)$ by simulating BM paths with the analytical metric tensor. The derivations of the analytical metric and parameterisation of the Swiss roll are shown in Appendix \ref{ax:swisspar}. Let $s_0$ with $radius=6$ and $width=3$ be the starting point of the BM. $N_{BM}=20000$ BM paths are simulated. The BM transition density is evaluated using \eqref{k:heat} at twenty nine target points $\{ s_j \in \M \subset \R^3| j \in \{1, \cdots,29 \} \}$ in the observation space. These target points are centred on $s_0$ and equally spaced. The diffusion time is fixed at 50. The results are plotted as the red solid line in Fig. \ref{fig:swiss-gl-kern}. The horizontal axis is the radius of the Swiss roll and the vertical axis is the transition density. The red density plot is asymmetric. When the radius is large, the transition density estimate decreases more quickly. 

If the geometry of the Swiss roll is unknown,  Bayesian GPLVM and GPLVM can be applied to learn the metrics from 250 grid points on the Swiss roll. The derivations of B-GPLVM metrics and GPLVM metrics are given in section \ref{bgplvm} and \ref{gplvm}. The BM trajectories are simulated using the estimated metric tensors. The heat kernel estimates using B-GPLVM metrics are plotted as the green dashed line in Fig.\ref{fig:swiss-gl-kern}. It is clear that the B-GPLVM results are very close to the red solid line.
The estimates using GPLVM metrics are plotted in brown dashed line. It is also close to the solid red line, but not as good as the B-GPLVM estimates. GPLVM performs point estimate of $x$ in the latent space while B-GPLVM estimates a Bayesian posterior of $x$. As a result, B-GPLVM is more robust in terms of estimating the latent variables \citep{damianou2016}. 
%The numerical comparison are given in Appendix??.

The heat kernel estimates from the Graph Laplacian (GL) approach (\cite{dunson2020diffusion}) in different data regime are also provided in Fig. \ref{fig:swiss-gl-kern}. When the number of points on the Swiss roll is 250, the GL kernel estimates are plotted as the blue dashed line. It is far from the solid red line and do not match the overall pattern of the analytical kernel estimates. When the number of grid points is increased to 1000 and 10000, the GL kernel estimates are plotted as the purple dashed line and black dash dotted line. The estimates are closer to the solid red line as the number of grid points increases. Comparing to the GL approach, the kernel estimates using B-GPLVM metrics achieve the best performance with much fewer points on the Swiss roll. B-GPLVM is used in the simulation study and real data applications in later sections.

\begin{algorithm}[tb]
   \caption{ Simulating BM sample paths on $\M$ for estimating $K_{heat}^t$ }
   \label{alg:bm}
\begin{algorithmic}
  \STATE Learn the metric $\G$ from the point cloud $\S = \{  s_i | i = 1,\ldots,n+v\ \} $. \COMMENT {\footnotesize use eqn \eqref{eqn:exmetric}  }
   \STATE  1.1 Generate BM trajectories on implicit manifold.
   \FOR{$i = 1,\ldots,n$   \COMMENT{ \footnotesize $n$ is the size of data points } }
     \FOR{$j = 1,\ldots,N_{BM}$   \COMMENT{\footnotesize $N_{BM}$ is No. of trajectories }  }
    \FOR{$ l = 1,\ldots,N_t$   \COMMENT{\footnotesize $N_t$ steps Brownian motion, $N_t\times \Delta t$  $\rightarrow$ max diffusion time} }
  % \IF{$x_i > x_{i+1}$}
 \STATE {\bf do}   \COMMENT{\footnotesize keep proposing x until it falls inside of the boundary }
   \STATE $q\left(x_{i,j}(l)|x_{i,j}(l-1)\right) = \mathbb N \left( x_{i,j}(l) |  \mu \left( x_{i,j}\left(l-1\right),\Delta t \right)  , \Delta t \G^{-1} \right )$  \COMMENT{ \footnotesize  use eqn \eqref{eqn:disProp} } 
\STATE {\bf While} $\V ar( \phi(x_{i,j})  ) > \alpha$, $x_{i,j}$ is outside of the boundary $\partial \M$.   \COMMENT{ \footnotesize use eqn \eqref{eqn:varb}  } 
   %\ENDIF
    \ENDFOR
     \ENDFOR
    \ENDFOR
    \RETURN $\bm x$
\STATE 1.2 Given a discrete choice of the diffusion time $t\in \{ \Delta t, 2\Delta t,\cdot \cdot \cdot, N_t\Delta t \}$, the covariance matrix $\Sigma^{t}$ is estimated based on the BM simulation from Algorithm 1.1.
\FOR{$i = 1,\ldots,n$} 
\FOR{$j = 1,\ldots,n$} 
\STATE $N_{\mathbf{A}_j}$ = which( x(t) $\in$ $\mathbf{A}_j$ )  \COMMENT{ {\small counting how many BM paths reach $\mathbf{A}_j$}  }
\STATE $K_{heat}^t(s_i,s_j)= \frac{ N_{\mathbf{A}_j}}{N_{BM}*V( \mathbf{A}_j)}$  \COMMENT{ {\small use eqn \eqref{k:heat} } }
\STATE $\Sigma_{ij}^t = \sigma_h^2 K_{heat}^t(s_i,s_j)$ 
\ENDFOR
\ENDFOR
\RETURN $\Sigma^t$
\end{algorithmic}
\end{algorithm}

From  \eqref{k:heat} we can see the construction of GPUM and the heat kernel requires simulating BM sample paths at each data point. Although the BM
simulations are trivially parallelizable, the computational cost can be high when the
number of data points is large. \cite{niu2019} proposed the sparse intrinsic GP on known manifold by introducing some inducing points. The number of inducing
points is much smaller than the number of data points. BM paths only need to be
simulated starting at the inducing points instead of every data point. The inducing point approximation summarizes the training data into a small set of inducing points, so that
inference could be done more efficiently \citep{QuioneroCandelaRasmussen2005}. Similar approach can be applied in the
GPUM when the manifold is unknown. Small number of inducing
points can be introduced in the learned latent space. The focus of this paper has been on developing the GPUM. The development of the sparse GPUM is for the future research.

\subsection{Optimising the kernel hyperparameters} \label{sec:optimpar}
Given a diffusion time $t$, we can generate the covariance matrix $\Sigma_{ {\text {\bf ff}} }^{t}$ for training data $\mathcal{D}$ using Algorithm \ref{alg:bm}. $\Sigma_{ \text{\bf ff} }^t$ can be obtained as follows: with the $i_{th}$ data point as the starting point, $N_{BM}$ trajectories are simulated to generate the $i_{th}$ row of $\Sigma_{ \text{\bf ff} }^t$. For each element of $\Sigma_{ \text{\bf ff}}^t$, $\hat{K}^t(s_i,s_j)$ is then estimated using  \eqref{k:heat}. The hyperparameters can be obtained by maximizing the log of the marginal likelihood  (over $f$) in \eqref{loglike}. %The log marginal likelihood function (over $f$) is given by \cite{Rasmussen2004}. 
The maximum BM diffusion time is set as $N_t \times \Delta t$. $\Delta t$ is the BM simulation time step as defined in \eqref{eqn:disProp}. $N_t$ is the number of simulation steps. $N_t$ covariance matrices $\Sigma_{ {\text {\bf ff}} }^{1 \ldots N_t}$ can be generated based on the BM simulations. Optimisation of the diffusion time $t$ can be done by selecting the corresponding $\Sigma_{ {\text {\bf ff}} }^{t}$ which maximizes the log marginal likelihood. Estimation of $\sigma_h$ follows standard optimisation routines, such as quasi-Newton.
\begin{align}
\label{loglike}
\log p( \bm{y}| s) &= \log \int p( \bm{y}| {\text {\bf f} }) p({\text {\bf f} }|s) d{\text {\bf f} }   \nonumber \\
&= -\frac{1}{2} \bm y^T (\Sigma_{ {\text {\bf ff}} }^{t} + \sigma_{noise}^2 I)^{-1} \bm y - \frac{1}{2} \log|\Sigma_{ {\text {\bf ff}} }^t +\sigma_{noise}^2I | - \frac{n}{2}\log2\pi.
\end{align}

Let ${\bf f}_*$ be a vector of values of $f(\cdot)$ at the unlabeled points in $\mathcal{V}$. Under the regression model in \eqref{eqn:reg} and the \IGP prior, we have the joint distribution of ${\bf y} $ and ${\bf f}_*$:
%The joint distribution of ${\bf f} $ and ${\bf f}_*$ is:
\begin{align}
p( {\text {\bf y}}, {\text {\bf f}}_*) = \mathcal{N} \left ( 0,
 \left [ \begin{array}{cc}  
 \Sigma_{{\text {\bf f}} {\text {\bf f}}} + \sigma_{noise}^2 I_n &\Sigma_{ {\text {\bf f}}  {\text {\bf f}}_*} \\[0.3em]
 \Sigma_{ {\text {\bf f}}_* {\text {\bf f}}} &\Sigma_{ {\text {\bf f}}_* {\text {\bf f}}_*} \\[0.3em]
                \end{array} \right ]
                \right ),
\end{align}
where $\Sigma_{{\text {\bf f}}_*{\text {\bf f}}}$ is the covariance matrix for training(labeled) and unlabeled data points. The predictive distribution is derived by marginalising out {\bf f}:
\begin{align}
p( {\text {\bf f}}_*| \bm y) = \int p( {\text {\bf f}}_*{\text {\bf f}}| \bm y) d{\text {\bf f}} = \mathcal N \left ( \Sigma_{ {\text {\bf f}}_*{\text {\bf f}}}  \left(\Sigma_{{\text {\bf f}}{\text {\bf f}}}  + \sigma_{noise}^2 I \right)^{-1}\bm y , \ \  \Sigma_{{\text {\bf f}}_*{\text {\bf f}}_*}-\left(\Sigma_{{\text {\bf f}}{\text {\bf f}}} +\sigma_{noise}^2 I \right)^{-1} \Sigma_{{\text {\bf f}}{\text {\bf f}}_*} \right).%\nonumber
\end{align}

%\newpage

\section{Simulation study on Swiss roll} \label{sec:simdata}

\begin{figure}[H]
    \centering
\subfigure[Swiss roll point cloud in $\R^3$ ]{\label{swiss:point3d}  \includegraphics[width=0.44\textwidth,height=0.3\textwidth]{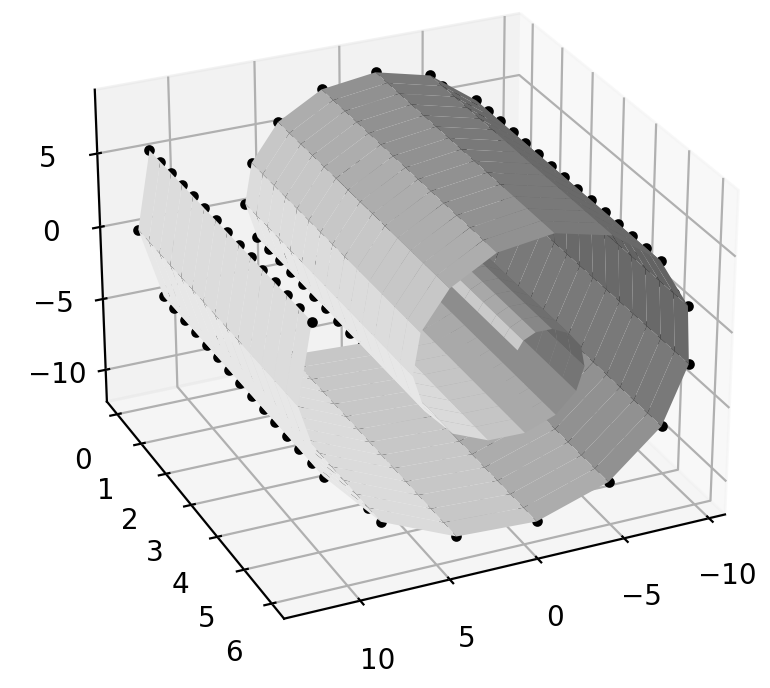} }
\subfigure[ Latent space and variance of mapping]{ \label{swiss:var} \includegraphics[width=0.45\textwidth,height=0.3\textwidth]{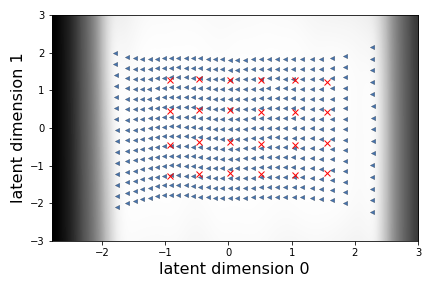} }
\subfigure[Latent space and magnification factor]{ \label{swiss:mag} \includegraphics[width=0.45\textwidth,height=0.3\textwidth]{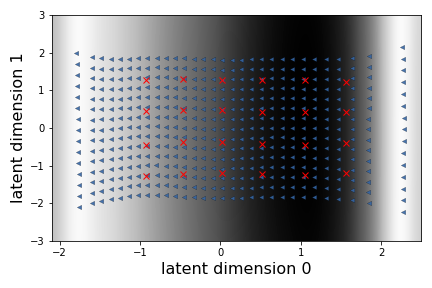} }
\subfigure[Boundary]{ \label{swiss:boud} \includegraphics[width=0.49\textwidth,height=0.35\textwidth]{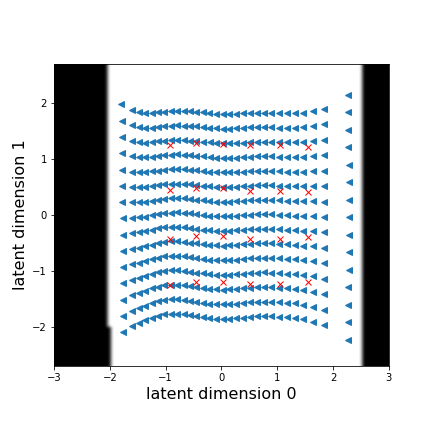} }
    \caption{ \label{fig:swissDR}
    \footnotesize{ (a) The Swiss roll represented as a point cloud in $\R^3$. (b) The latent space constructed from Bayesian GPLVM. The blue triangles are unlabeled points and red crosses are labeled points. The background color represents the variance of the mapping, a dark background represents high uncertainty. (c) The same latent space visualization with the background color representing the magnification factor, a dark background represents high magnification factor. (d) The same latent space visualization highlighting the boundary. The dark region is outside of the boundary and the white region is inside of the boundary.}   
  }
\end{figure}

 In this section, we carry out a simulation study for a regression model with synthetic data on the Swiss roll which is a two dimensional manifold depicted by a point cloud in $\R^3$. The point cloud is plotted in Fig. \ref{swiss:point3d}. It is comprised of the set of labeled points $n=24$ and the set of unlabeled points $v=450$. Both labeled and unlabeled observed points are used in B-GPLVM to learn the latent space. The point cloud is unfolded into a flat surface as in Fig. \ref{swiss:var}. The unlabeled points are plotted as blue triangles and the labeled points are in red in the latent space. The variance of the mapping is plotted as the gray background of the latent space. It is clear from Fig. \ref{swiss:var} that the background color gets darker in the region further away from the blue triangles. This indicates the variance of the mapping %$Var(\phi(x))$ 
gets bigger in the region which is far from the observations. %when $x$ is further away from the observed latent points.
 
 Based on the definition in \eqref{eqn:bound}, the boundary $\partial \M$ is plotted in Fig. \ref{swiss:boud}. The black regions on the left and right sides of Fig. \ref{swiss:boud} are outside of $\partial \M$. The white area in the middle is within $\partial \M$. The magnification factor is plotted as the background color in Fig. \ref{swiss:mag}. Similar to Fig. \ref{fig:BM}, the horizontal axis can be interpreted as the scaled radius of the Swiss roll.  
 When the radius is bigger, the corresponding magnification factor is larger (the color is darker).  However, as there are fewer data points available at the tail of the Swiss roll, the estimation of the implicit manifold from B-GPLVM becomes less accurate. This can be observed on the right end of Fig. \ref{swiss:mag} where the last column of the observed latent points is further away from the rest. Once the metric $\mathcal{G}$ and the boundary $\partial \M$ are learned from B-GPLVM, we can estimate the heat kernel by simulating BM paths on the implicit manifold. An example of a BM sample path on Swiss roll is shown in Fig. \ref{fig:BM}.

\begin{figure}[H]
    \centering
    \subfigure[True function and data points]{\label{swiss:true}  \includegraphics[width=0.45\textwidth,height=0.4\textwidth]{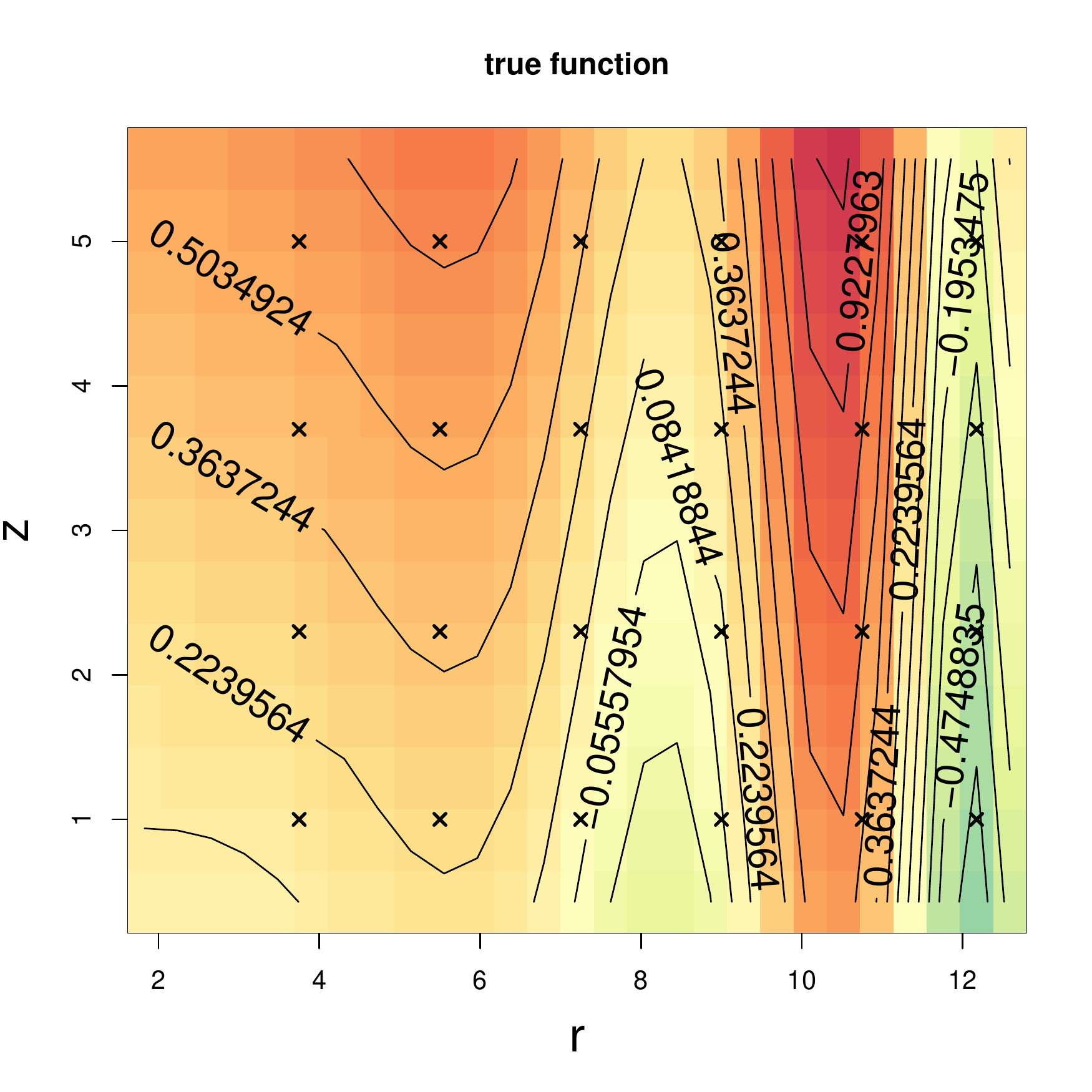}}
     \subfigure[ \small{$\R^3$ GP  prediction }]{\label{swiss:3d}  \includegraphics[width=0.45\textwidth,height=0.4\textwidth]{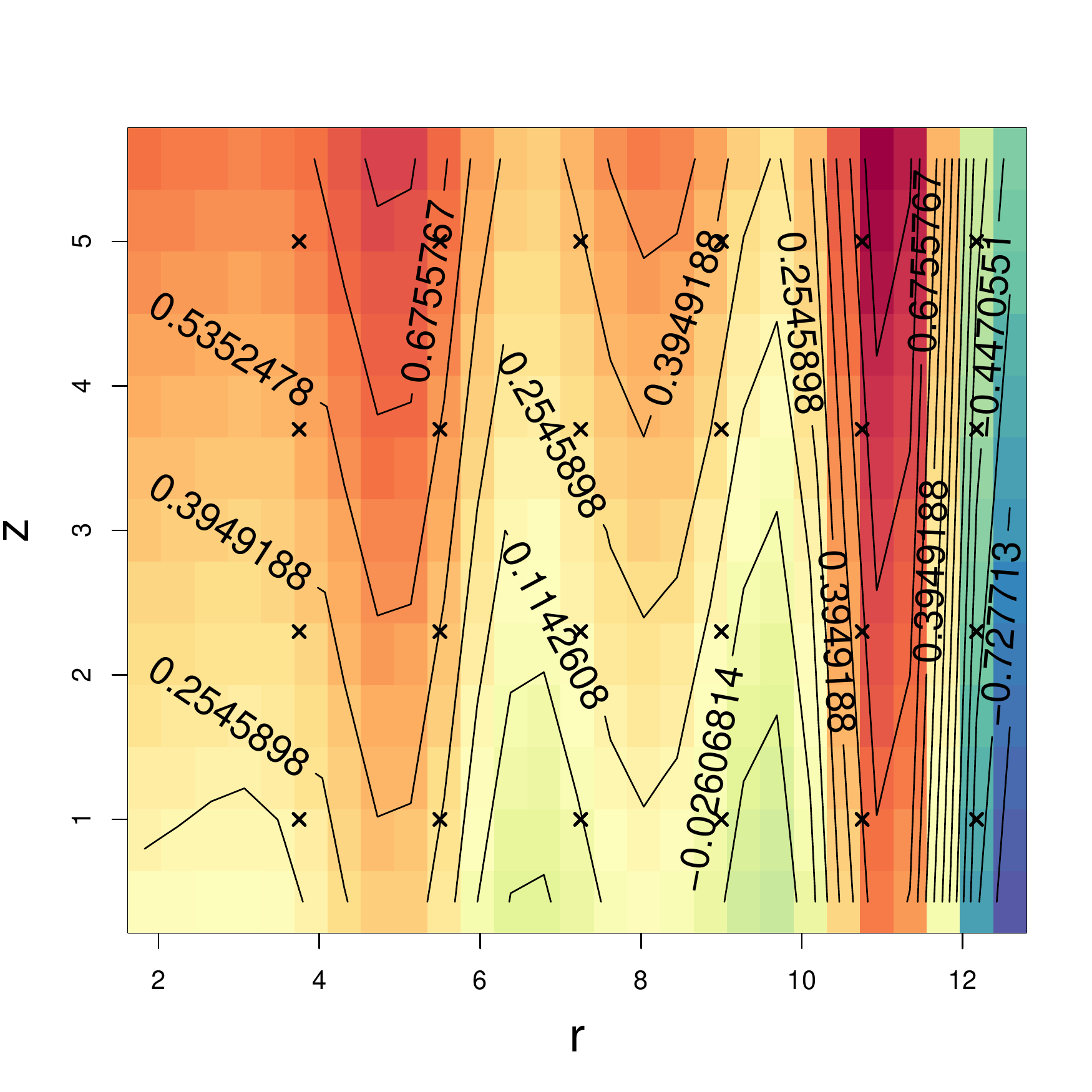}}
    \subfigure[\IGP prediction]{\label{swiss:gp}  \includegraphics[width=0.45\textwidth,height=0.4\textwidth]{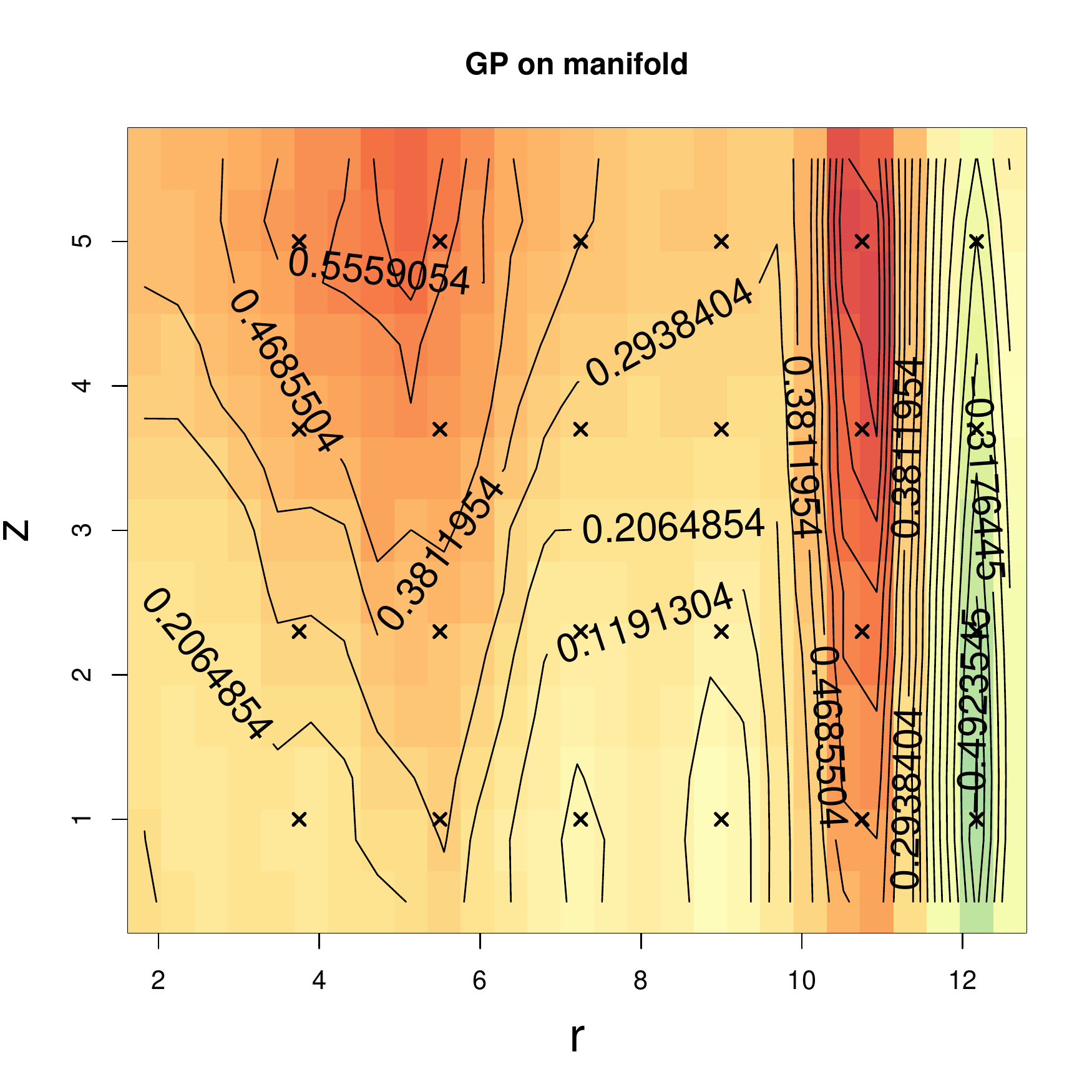}}
    \subfigure[ Comparison of predictions when fixing z ]{ \label{swiss:compare1d} \includegraphics[width=0.45\textwidth,height=0.4\textwidth]{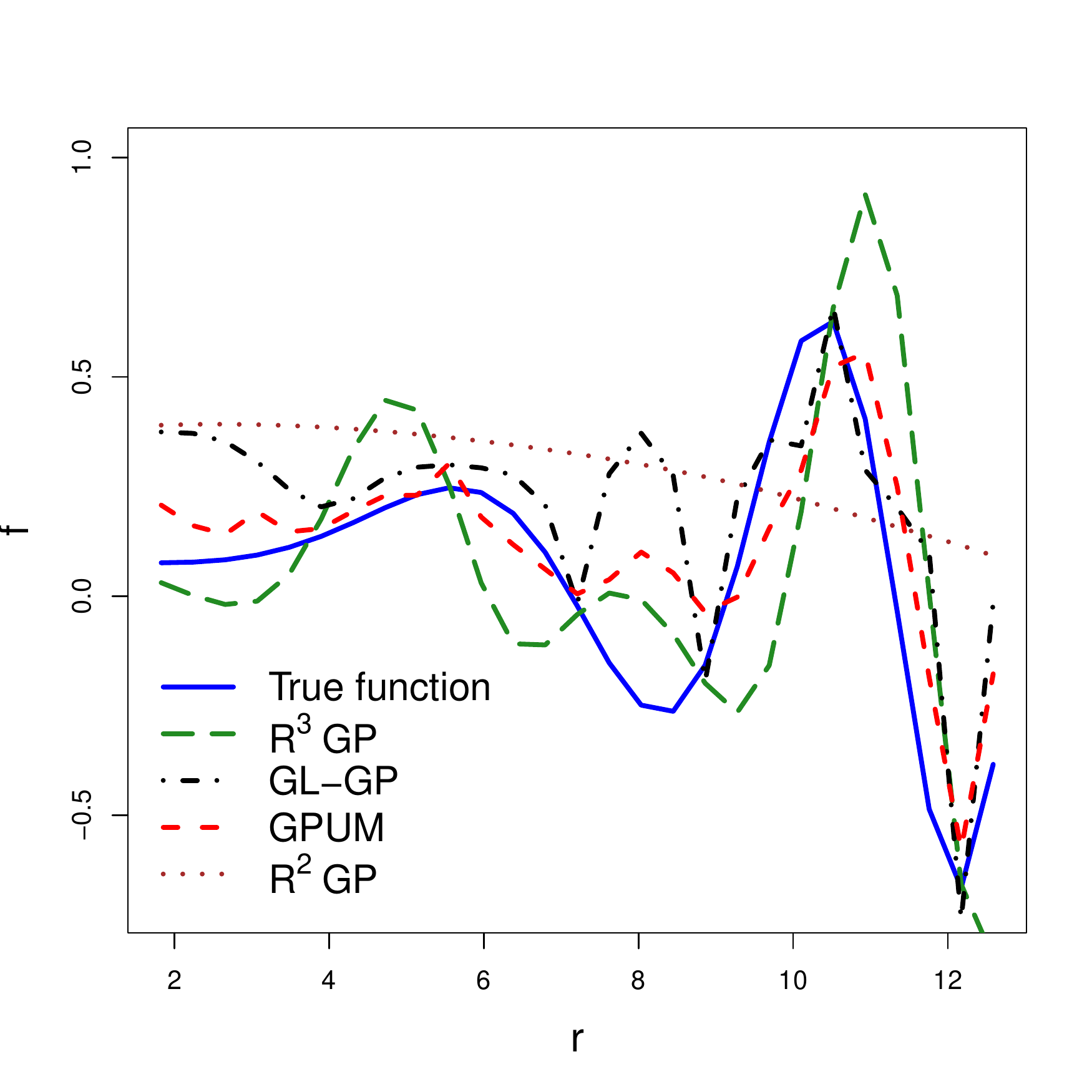}}%swissGP2-2.pdf. $\R^2$ GP prediction
    \caption{\label{fig:swiss2d}
   { \footnotesize
(a) The true function in the unfolded Swiss roll. The labeled points are marked with black crosses. (b) The prediction of $\R^3$ GP. $\R^3$ GP is constructed using the Euclidean distance in $\R^3$. (c) The prediction of \IGP. (d) Comparing the predictions of all methods on a spiral in the swiss roll by fixing the z coordinate at 4 and changing the radius from 2 to 12.5. The vertical axis represent the value of the prediction. The horizontal axis represents the radius. %(d) The prediction of $\R^2$ GP. $\R^2$ GP is constructed using the $\R^2$ Euclidean distance in the latent space. The geometric properties such as metric tensor and boundary are ignored in $\R^2$ GP.
    }  
 }
\end{figure}

For the labeled points, the response variables are 
 \begin{align*}
y_i = f(s_i^1,s_i^2,s_i^3) + \epsilon_i, \ \ i=1\ldots n, \ \ s_i\in \R^3
 \end{align*}
 
where $f$ is the unknown regression function, $s_i$ is the coordinate of the observed point in $\M \subset \R^3$. For better visualisation, the true function is plotted in the unfolded Swiss roll in Fig.\ref{swiss:true} using a two dimensional analytical parameterisation. The coordinates are radius and width. 24 labeled observations are marked as black crosses. The true function values are indicated by the color codes and contours. The regression function varies slowly when the radius is small and rapidly when the radius is big.

We first apply the Euclidean $\R^3$ GP (the standard GP as in \cite{Rasmussen2006} Chapter 4) constructed by the squared exponential kernel with $\R^3$ Euclidean distance. With this kernel setting, the $\R^3$ GP completely ignores the interior structure of the manifold and lets the inner layer and outer layer of the Swiss roll interact. The predictive means on the unlabeled points are shown in Fig. \ref{swiss:3d}. Compared to the true function in Fig.\ref{swiss:true}, the overall shape of $\R^3$ GP prediction contours is more wiggly. The color coding of Fig.\ref{swiss:3d} is also very different in the regions where the radius (horizontal axis) is 6, 8 and 10. In the second model, the $\R^2$ Euclidean distance in the latent space is used in the squared exponential kernel to construct the $\R^2$ GP. The geometric properties such as the metric and magnification factors are ignored. The color coding of the predictive means for $\R^2$ GP is shown in Fig. \ref{swiss:r2} in Appendix \ref{ax:swisspar}. Since the regression function is nonstationary in the latent space, the $\R^2$ GP is underfitting. The \IGP predictive mean is shown in Fig. \ref{swiss:gp}. The overall pattern of the  \IGP prediction is similar to the true function. The shape of the contours and the color coding of Fig.\ref{swiss:gp} are consistent with Fig.\ref{swiss:true}. The prediction of the GL-GP is shown Fig.\ref{swiss:gl} in Appendix \ref{ax:swisspar}. Since the GL kernel estimates in Fig. \ref{fig:swiss-gl-kern} are far from the analytical kernel, the GL-GP prediction is also poor. A one dimensional comparison is generated by plotting the predictions of all methods at $z=4$ and varying the radius from 2 to 12.5 in Fig.\ref{swiss:compare1d}. It is clear the $\R^2$ prediction(brown dotted line) is under fitting. The $\R^3$ prediction(green dashed line) is oscillating in the opposite direction of the ground truth (blue solid line). The GL-GP prediction (black dashed line) is also oscillating around the blue solid line. The \IGP prediction (red dashed line) achieves the best performance and follows the overall pattern of the ground truth.

In order to evaluate the performance of different methods in different data regimes, we consider three scenarios with different numbers of unlabeled grid points on the Swiss roll from $v=250$ to $v=450$ and $v=800$. These unlabeled grid points are used to estimate the manifold for \IGP, GM-GP, GL-GP and are also used as the test set of the regression accuracy. We constructed twenty sets of training points by randomly selecting $n=23$ labeled points from the labeled set. For each training set, \IGP, GM-GP, GL-GP, $\R^2$GP and $\R^3$GP have been applied to make predictions at the points in the test set. The root mean square errors (RMSE) are calculated between the true function values and the predictive means of all five models. The mean and standard deviation of the root mean square errors are reported in Table \ref{tb:swiss}. GPUM’s results are consistent
across all three scenarios and significantly better than all other methods. The GL-GP and GM-GP have similar performance. Both models perform better when the number of the grid points is large. They are significantly better than the Euclidean GPs. Comparing to the graph based approaches, GPUM achieves the minimum mean RMSE with fewer points on the manifold.

%The numerical comparison is in Table \ref{tb:swiss}. The mean  are compared in these three scenarios.  
%The \IGP outperforms the other two models and achieves the minimum RMSE 0.158.
%we have applied GPUM, GM-GP, GL-GP and Euclidean GP in the Swiss roll. 

\begin{table}[h]
\caption{\label{tb:swiss}Comparison of the root mean squared errors of five methods on Swiss roll. Values in parentheses show the standard deviation.}
%\vskip 0.15in
%\vskip 0.1in
\begin{center}
\begin{small}
\begin{sc}
\begin{tabular}{lccccr}
\hline
    &   $\R^3 GP$ & $\R^2 GP$ &  \IGP & GL-GP & GM-GP\\
\hline
mean RMSE $v=250$ &0.284(0.006) & 0.293(0.005)&0.163(0.020)& 0.243(0.003)&0.231(0.001) \\
mean RMSE $v=450$ & 0.298(0.007)  & 0.290(0.005) & 0.162(0.003) &0.220(0.002) & 0.207(0.002)\\
mean RMSE $v=800$ &0.287(0.006)&0.282(0.005)& 0.164(0.002)&0.216(0.001) & 0.206(0.001)\\
\hline
\end{tabular}
\end{sc}
\end{small}
\end{center}
\vskip -0.1in
\end{table}

% 0.437(0.218)
% 0.507(0.333)

\begin{figure}[H]
    \centering
\subfigure[ Latent space and uncertainty of mapping ]{\label{wifi:latent}  \includegraphics[width=0.44\textwidth,height=0.3\textwidth]{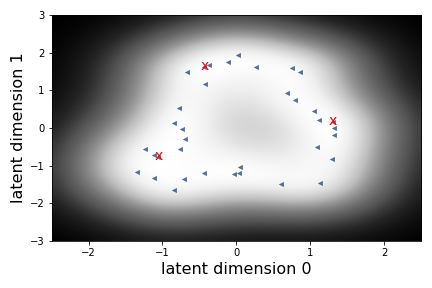} }
\subfigure[ Latent space and boundary]{ \label{wifi:bound} \includegraphics[width=0.51\textwidth,height=0.33\textwidth]{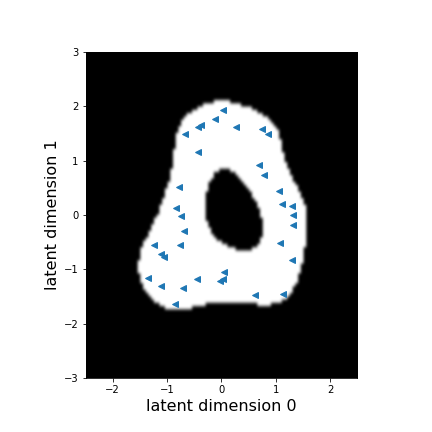} }
\subfigure[Latent space and magnification factor]{ \label{wifi:mag} \includegraphics[width=0.45\textwidth,height=0.3\textwidth]{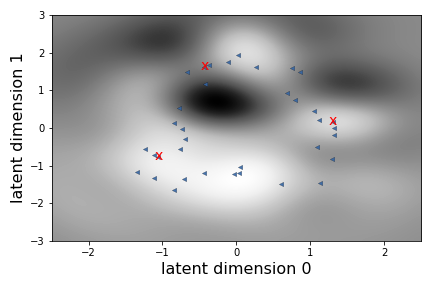} }
\subfigure[A BM trajectory]{ \label{wifi:bm} \includegraphics[width=0.45\textwidth,height=0.3\textwidth]{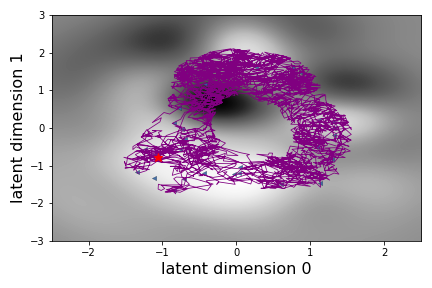} }
    \caption{ \label{fig:wifi}
    \footnotesize{ (a)
The latent space constructed from Bayesian GPLVM. The blue triangles are unlabeled points and the red crosses are labeled points. The background color represents the variance of the mapping, a dark background represents high uncertainty. (b) The same latent space visualization highlighting the boundary. The dark region is outside of the boundary and the white region is inside of the boundary. (c) The same latent space visualization with the background color representing the magnification factor, a dark background represents a high magnification factor. (d) A BM trajectory is shown in purple. The red star is the starting location.}   
  }
\end{figure}

\section{Location estimation from WiFi signal} \label{sec:wifi}

Indoor location estimation has tremendous value but standard location estimation techniques such as Global Positioning System (GPS) do not work indoors. Instead, indoor wireless signals from devices such as WiFi access points can be exploited for location estimation. In this section, we consider the problem of indoor 2D location estimation from WiFi access point signal strengths. We use the WiFi data collected by \cite{ferris2007}, in which a series of WiFi signal strength traces are collected by a mobile device which travels in a one floor university building. The 2D location coordinates of the mobile device are also recorded by a click-to-map based annotation program. The total number of WiFi access points in this dataset is 30. As it is often expensive to collect labeled data, we mimic a low data indoor location estimation scenario by assuming that the true locations of only three points are known and we aim at predicting the indoor locations of the remaining points based on the WiFi signals. 

This location estimation problem can be treated as a regression problem, in which the location coordinate $y$ of the mobile device can be modelled as a function of the high dimensional WiFi signal
\begin{align*}
y_i = f(s_i) + \epsilon_i, \ \ i=1\ldots n, \ \ s_i\in \R^{30}
\end{align*}
where $f$ is the unknown regression function and WiFi signal $s_i$ is represented by a 30 dimensional vector. Here we consider the WiFi signal measurements at $n+v=36$ locations. Only $n=3$ of the locations are labeled with one dimensional location coordinates of the mobile device. To avoid selecting the three points clustered together, we randomly pick the points from three different regions respectively, where the union of the regions cover the whole dataset. Different methods are applied to estimate the coordinates of the mobile device in the testing sets (the unlabeled $v=33$ locations). Twenty training and testing sets are generated from this random selection.

Unlike the simulation study of the Swiss roll, we cannot plot the high dimensional point cloud of the WiFi signals. The implicit manifold is also unknown. We first estimate a $q=2$ dimensional latent space (the chart of the underlying manifold) using B-GPLVM. The value of $q$ is determined by the ARD contributions which measure how much each dimension is contributing to the latent space. The input dimensions are sorted based on the relevance assigned by the scaling in B-GPLVM. The plot of ARD contributions is shown in Appendix \ref{ax:wifi}. The latent space is plotted with the variance of the mapping as the gray background in Fig. \ref{wifi:latent}. The 36  WiFi signal strength measurements are represented by the blue triangles. The training set (the labeled points) is marked by the red crosses. The dark color value is for high uncertainty. Since the mobile device moves in a loop closure, the latent points forms a closed loop in the latent space. The boundary of the implicit manifold $\partial M$ is shown in Fig. \ref{wifi:bound} which is defined by \eqref{eqn:bound}. The magnification factor is  plotted as the gray background in Fig.\ref{wifi:mag}. The dark color value is for high magnification factor. A sample path of BM in the implicit manifold is plotted in Fig. \ref{wifi:bm}. With the Neumann boundary condition the BM path can only exist within the boundary.

\begin{figure}[t]
    \centering
    \subfigure[Ground truth]{\label{wifi:true}  \includegraphics[width=0.315\textwidth,height=0.3\textwidth]{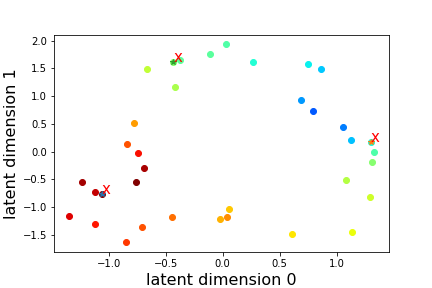}}
     \subfigure[ \small{ $\R^{30}$ GP prediction }]{\label{wifi:hd}  \includegraphics[width=0.315\textwidth,height=0.3\textwidth]{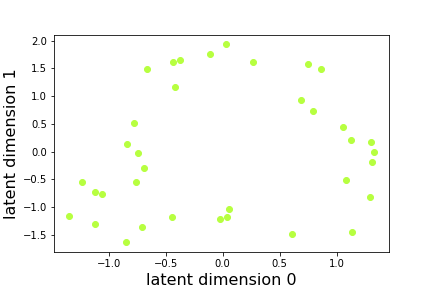}}
    \subfigure[\IGP  prediction]{\label{wifi:in-GP}  \includegraphics[width=0.315\textwidth,height=0.3\textwidth]{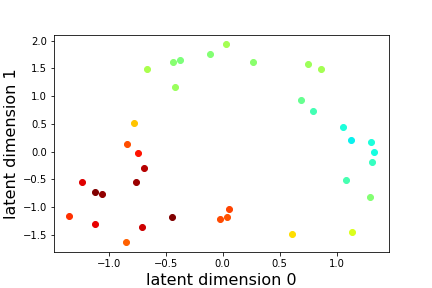}}
   % \subfigure[$\R^2$ GP prediction]{ \label{coil:2d} \includegraphics[width=0.35\textwidth,height=0.3\textwidth]{coil2D.png}}
    \caption{\label{fig:wifiGP}
   {\footnotesize
    Comparison of \IGP and Euclidean GPs in WiFi signal example. (a) The ground truth is plotted with color at the observed points in the latent space. (b) $\R^{30}$ GP prediction. $\R^{30}$ GP is constructed using the Euclidean distance of WiFi signals in $\R^{30}$. (c) \IGP prediction. 
    }  
 }
\end{figure}

The coordinates of the mobile device are estimated by five different methods. The ground truth values are marked by different colors in the latent space in Fig. \ref{wifi:true}. If we ignore the interior structure of the manifold, a $\R^{30}$ GP using the squared exponential kernel with $\R^{30}$ Euclidean distance of the WiFi signals is applied. The predictive means of the $\R^{30}$ GP are plotted in Fig.\ref{wifi:hd}. It is clear the $\R^{30}$ GP prediction is very poor. The color coding of the prediction is different from the ground truth. The \IGP  predictive mean is shown in Fig. \ref{wifi:in-GP}. The overall pattern of the \IGP prediction is similar to the ground truth. In the third case, the $\R^2$ Euclidean distance in the latent space is used with the squared exponential kernel to construct the $\R^2$ GP. The prediction results are shown in Fig.\ref{fig:wifi2d} in Appendix \ref{ax:wifi}. The Graph Laplacian based approaches such as GL-GP and GM-GP have also been applied. Since the number of the observations are relatively low ($n+v=36$), the graph based methods result in poor approximation of the implicit manifold. The GL-GP's performance is similar to the $\R^{30}$GP. The mean and standard deviation of the root mean square errors of the twenty testing sets are also calculated in Table \ref{tb:wifi} for all five models. \IGP significantly outperforms the other four models and achieves the minimum mean RMSE.

\begin{table}[t]
\caption{\label{tb:wifi} Comparison of the root mean squared errors of five methods on WiFi signal data. Values in parentheses show the standard deviation. }
\vskip 0.15in
\begin{center}
\begin{small}
\begin{sc}
\begin{tabular}{lccccr}
\hline
    & $\R^{30}$GP & $\R^2$GP & \IGP &  GL-GP &  GM-GP \\
\hline
mean RMSE & 5.57(1.43)  & 4.83(1.83) & 4.11(0.88) & 5.6(1.15) & 6.04 (0.57) \\
%Standard Deviation & 0.57 & 1.15 & 1.43  &1.83  & 0.88 \\
\hline
\end{tabular}
\end{sc}
\end{small}
\end{center}
\vskip -0.1in
\end{table}
% 6.41 sd 0.48

\section{Camera angle estimation from images} \label{sec:real}
\begin{figure}[t]
    \centering
\subfigure[ COIL images ]{\label{coil:image} \includegraphics[width=0.31\textwidth,height=0.28\textwidth]{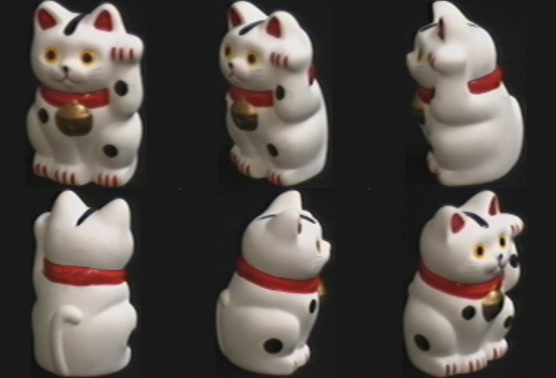} }
\subfigure[ Latent space and variance of mapping]{ \label{coil:var} \includegraphics[width=0.31\textwidth,height=0.3\textwidth]{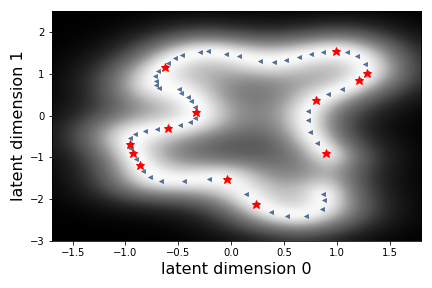} }
\subfigure[Latent space and magnification factor]{ \label{coil:mag} \includegraphics[width=0.31\textwidth,height=0.3\textwidth]{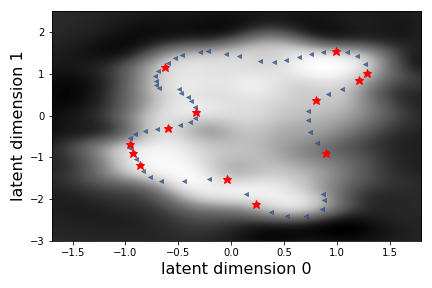} }
    \caption{ \label{fig:coil}
    \footnotesize{ (a) Examples of six COIL object images. (b) The latent space constructed from Bayesian GPLVM. The blue triangles are unlabeled points and red stars are labeled points. The background color represents the variance of the mapping. The dark color represents high uncertainty. (c) The same latent space visualization with the background color representing the magnification factor. The dark color represents a high magnification factor.
      }   
  }
\end{figure}

In this section, we consider the problem of estimating the camera angle from images. We consider the setting in which an object is placed on a turntable and a set of images are taken at different angles with respect to the camera. We aim at recovering the camera angles associated with individual images by knowing the true camera angles for some of the images. We use the images from the COIL data set \citep{nene1996}. Here we consider 66 images of object-14 (a toy cat). The camera angles range from 15 degrees to 340 degrees. 
The raw images are converted to grayscale and downscaled to $32 \times 32$. 
The raw pixels of each image are flattened into a 1024 dimensional vector. 
The dimension of the original image space is $p=1024$. Six image examples are given in Fig. \ref{coil:image}. We estimate a $q= 2$ dimensional latent space using Bayesian GPLVM. The two dimensional latent space is plotted with the variance of the mapping as the gray background in Fig. \ref{coil:var}. A dark background represents a high uncertainty in the corresponding region. The unlabeled data points are marked as blue triangles and the labeled data points as red stars. The overall shape of the latent points in Fig.\ref{coil:var} looks like a ring with a gap in the lower right. The magnification factor is plotted as the gray background in Fig. \ref{coil:mag}. A dark background represents a high magnification factor. The boundary of the implicit manifold $\partial \M$ is shown in Fig.\ref{coil:bound} in Appendix \ref{ax:coil}. %Fig. \ref{coil:bound}. %It is defined by \eqref{eqn:bound} and the uncertainty of the mapping as in Fig. \ref{coil:var}. A sample path of BM in the implicit manifold is plotted in Fig.\ref{coil:BMpaths}.% The BM trajectory is simulated with a big diffusion time $t$ to demonstrate the effect of the boundary. %With the Neumann boundary condition the BM paths can only exist within the boundary. 

\begin{figure}[t]
    \centering
    \subfigure[True function]{\label{coil:true}  \includegraphics[width=0.315\textwidth,height=0.3\textwidth]{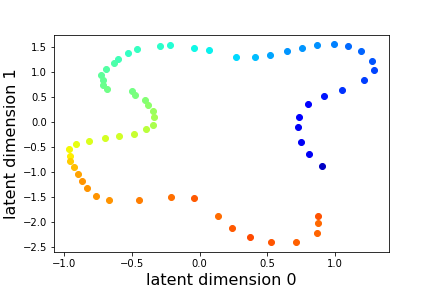}}
     \subfigure[ \small{ $\R^{1024}$ GP prediction }]{\label{coil:hd}  \includegraphics[width=0.315\textwidth,height=0.3\textwidth]{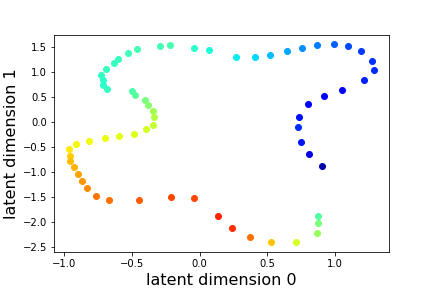}}
    \subfigure[\IGP on manifold prediction]{\label{coil:in-GP}  \includegraphics[width=0.315\textwidth,height=0.3\textwidth]{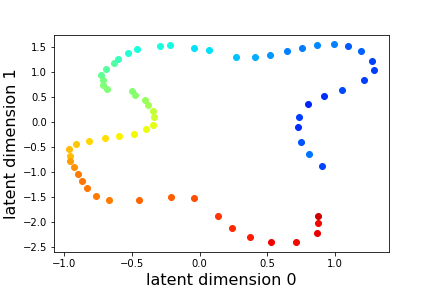}}
   % \subfigure[$\R^2$ GP prediction]{ \label{coil:2d} \includegraphics[width=0.35\textwidth,height=0.3\textwidth]{coil2D.png}}
    \caption{\label{fig:coilGP}
   { \footnotesize
     Comparison of \IGP and Euclidean GPs in COIL images example. (a) The true function is plotted with color at the observed points in the latent space. (b) $\R^{1024}$ GP prediction. $\R^{1024}$ GP is constructed using the Euclidean distance of data points in $\R^{1024}$. (c) \IGP prediction. 
    }  
 }
\end{figure}

The scaled camera angle $y$ is modelled as a function of the image.
 \begin{align*}
y_i = f(s_i) + \epsilon_i, \ \ i=1\ldots n, \ \ s_i\in \R^{1024} .   
 \end{align*}
where $f$ is the unknown regression function, and $s_i$ is represented by a 1024 dimensional vector. n=13 images are randomly selected and used as the training set (labeled data). They are plotted as the red stars in the latent space in Fig.\ref{coil:var}. The remaining 53 images are used as the testing set (unlabeled data)  which are plotted as the blue triangles in the latent space in Fig. \ref{coil:var}. The true angle values are marked by different colors on the observed points in the latent space in Fig. \ref{coil:true}. Different methods have been applied to estimate the scaled camera angles. The $\R^{1024}$GP using the squared exponential kernel with $\R^{1024}$ Euclidean distance is applied to the image data first. The predictive means of the $\R^{1024}$GP are plotted in Fig.\ref{coil:hd}. The color coding of the prediction at the lower right of the plot is bright green which is different from the truth. Ignoring the boundary and the magnification factor in the latent space, the $\R^2$GP using the squared exponential kernel with $\R^2$ Euclidean distance in the latent space is applied to the image data. The predictive means are shown in  Fig.\ref{coil:2d} in Appendix \ref{ax:coil}. The predictive means of \IGP are plotted in Fig.\ref{coil:in-GP}. The overall pattern of the \IGP prediction is very similar to the true function. Since the boundary of the implicit manifold is defined in the lower right region of the latent space, the \IGP does not smooth across the boundary and gives a better prediction. Ten training and testing
sets are generated from the random selection. 
The mean and standard deviation of the root mean square errors are calculated for the ten testing sets for all methods in Table \ref{tb:coil}. The GPUM achieve the smallest mean RMSE. It is significantly better than  the GL-GP and GM-GP. The difference between the \IGP and the $\R^{1024}$GP is not significant. The boxplot of the root mean squared errors are shown in Fig.\ref{coil:box} in Appendix \ref{ax:coil}.

%It is clear that the \IGP outperforms the other four methods and the differences are significant.  
 
\begin{table}[t]
\caption{\label{tb:coil}Comparison of the root mean squared errors of five methods on COIL images. Values in parentheses show the standard deviation.}
\vskip 0.15in
\begin{center}
\begin{small}
\begin{sc}
\begin{tabular}{lccccr}
\hline
&$\R^{1024}$GP &$\R^2$GP  & \IGP &  GL-GP& GM-GP  \\
\hline
mean RMSE & 0.097(0.040) &0.094(0.042) & 0.060(0.038)  & 0.116(0.027) & 0.143(0.026) \\
%Standard Deviation &0.026 & 0.027 & 0.040  &0.042 & 0.038\\
%mad &0.031 & 0.033 & 0.045  &0.052 & 0.027\\
\hline
\end{tabular}
\end{sc}
\end{small}
\end{center}
\vskip -0.1in
\end{table}

\section{Discussion} \label{discuss}

Our work provides a novel framework for regression on implicit manifolds embedded in high dimensional point clouds. The geometry of the implicit manifold is learned by probabilistic latent variable models. This gives a distribution over a smoothly changing local metric at each point in the latent space.  The boundary of the implicit manifold is defined according to the uncertainty of the mapping. The expression for the expected local metric is derived and used to simulate BM sample paths on manifolds. We have proved the BM transition density estimation is coordinate independent in section \ref{sec:density}. This allows us to compare the transition density estimates from different parameterisations of the same manifold. The BM simulation using the B-GPLVM metric gives similar transition density estimation results as the BM simulation using the analytical metric. \IGP is constructed by employing the equivalence relationship between heat kernels and the transition density of BM on manifolds. This allows the \IGP to incorporate the intrinsic geometry of the implicit manifold for inference while respecting the interior constraints and boundary. The experiment results in section \ref{sec:simdata}, \ref{sec:wifi} and \ref{sec:real} indicate that \IGP achieves significant improvements over Euclidean GPs and Graph Laplacian based GPs. Although the simulation of BM paths can be easily parallelised, Algorithm \ref{alg:bm} can still be computational expensive if the sample size is large. The number of the sample paths can be reduced by leveraging the idea of pseudo data, which has been widely studied in the literature of sparse GPs \citep{QuioneroCandelaRasmussen2005,niu2019}.

\appendix
\section{ Proof of Theorem 1 and Corollary 2 \label{ax:independent} }
The SDE in (22) can be derived by expressing the heat equation in local coordinates as a Fokker-Planck equation, which in particular implies its coordinate independence. In this appendix, however, we will provide a direct verification that (22) is invariant under coordinate changes. %\ref{eqnBM}. \ref{eqnBM}

Some notations we need in the proof are summarised here. $\M$ is a manifold of dimension $q$ with a riemannian metric $g_{\mathbb{M}}$. $\theta: \mathbb{U}\rightarrow \mathbb{M}$ and $\bar{\theta}: \bar{\mathbb{U}} \rightarrow \mathbb{M}$ are two local coordinate charts of $\M$ where $\mathbb{U}$ and $\bar{ \mathbb{U} }$ are open subsets of $\R^q$. 
\begin{itemize}
\item $\psi=\bar{\theta}^{-1} \circ \theta: \mathbb{U} \rightarrow \bar{\mathbb{U}}$ is the change of coordinate.
$\psi^{-1}=\theta^{-1} \circ \bar{\theta}: \bar{\mathbb{U}} \rightarrow \mathbb{U}$ is the inverse change of coordinate. (assume $\theta \left( \mathbb{U} \right) = \bar{\theta}\left( \bar{\mathbb{U}} \right)$ without loss of generality.)
\item $\left(x^{1}, \ldots, x^{q}\right)$ denotes the standard coordinates in $\mathbb{U} \in \R^q$.
$\left(\bar{x}^{1}, \ldots, \bar{x}^{q}\right)$ denotes the standard coordinates in $\bar{\mathbb{U}} \in \R^q$.
\item $g$ denotes matrix representation of $g_{\mathbb{M}}$ in $\mathbb{U}$ (via $\theta$) i.e. $g_{i j}=g_{\mathbb{M}}\left(\theta_{*} \frac{\partial}{\partial x^{i}}, \theta_{*} \frac{\partial}{\partial x^{j}}\right)$. $g^{ij}$ denotes the element of $g^{-1}$. $G = det(g)$. Similarly for $\bar{g}$ , $\bar{g}_{ij}$, $\bar{g}^{ij}$, $\bar{G}$.
\item  $D \psi$  is the matrix derivative of  $\psi = \left(\psi^{1}\left(x^{1}, \ldots x^{q}\right), \ldots, \psi^{q}\left(x^{1}, \ldots, x^{q}\right)\right)$, i.e. $(D \psi)_{i}^{j}=\frac{\partial \psi^{j}}{\partial x^{i}} $. Using the chain rule  $\Rightarrow  D \psi^{-1} \cdot D \psi=I_{q}$. %or $ D \phi^{-1}=(D \phi)^{-1}$
\item $\frac{\partial}{\partial x^{i}}=\sum_{j=1}^{q} \frac{\partial \psi^{j}}{\partial x^{i}} \frac{\partial}{\partial \bar{x}^{j}}=\sum_{j=1}^{q}(D \psi)_{i}^{j} \frac{\partial}{\partial \bar{x}^{j}} \quad \Rightarrow \quad g_{i j}=(D \psi)_{i}^{k} \bar{g}_{k \ell}(D \psi)_{j}^{\ell} $ $ \quad \Rightarrow g^{-1}=\left(D \psi^{-1}\right)^{\top}  \bar{g}^{-1} D \psi^{-1}$,
$G=(\operatorname{det} \mathcal{D} \psi)^{2} \bar{G}$.
\end{itemize}

\begin{proof}
Consider a stochastic process in $\mathbb{U} \in \R^q$ defined by:
\begin{align}
d x^{i} &= \frac{1}{2} \frac{1}{\sqrt{ G }} \sum_{j=1}^{q} \frac{\partial}{\partial x^{j}}\left(\sqrt{ G } g^{i j}\right) dt + \left(g^{-\frac{1}{2}} d B\right)^{i}, \quad i=1, \ldots, q. 
\end{align}
Note that
\begin{align*}
\quad d x^{i} d x^{j}&=\left(g^{-\frac{1}{2}} d B\right)^{i}\left(g^{-\frac{1}{2}} d B\right)^{j} 
=\left( \sum_{k=1}^{q}\left(g^{-\frac{1}{2}}\right)^{i k} d B^{k}\right) \cdot\left(\sum_{l=1}^{q}\left(g^{-\frac{1}{2}}\right)^{j l} d B^{l}\right) \\ 
&=g^{i j} d t  \ \   \   i,j = 1,\ldots , q.
\end{align*}
$\psi$ maps the above process in $\mathbb{U}$ to a process in $\bar{\mathbb{U}}$ defined by:
\begin{align}
\label{apxeqn:coord}
d \bar{x}^{i} &=d \psi^{i}\left(x^1, \ldots, x^{q}\right) \nonumber \\
&=\sum_{j=1}^{q} \frac{\partial \psi^{i}}{\partial x^{j}} d x^{j}+\frac{1}{2} \sum_{j=1}^{q} \sum_{k=1}^{q} \frac{\partial^{2} \psi^{i}}{\partial x^{j} \partial x^{k}} d x^{j} d x^{k} \nonumber \\
&= \frac{1}{2} \sum_{j=1}^{q} \sum_{k=1}^{q}\left[(D \psi)_{j}^{i} \cdot \frac{1}{\sqrt{G}} \frac{\partial}{\partial x^{k}}\left(\sqrt{G} g^{jk} \right)+\frac{\partial^{2} \psi^{i}}{\partial x^{j} \partial x^{k}} g^{j k}\right] d t +\sum_{j=1}^{q}(D \psi)_{j}^{i}\left(g^{-\frac{1}{2}} d B\right)^{j}.
\end{align}
The $dt$ term in \eqref{apxeqn:coord} can be written as follows
\begin{align}
&\frac{1}{2} \sum_{j=1}^{q} \sum_{k=1}^{q} \left[ (D \psi)_{j}^{i} \cdot \frac{1}{det D \psi} \cdot \frac{1}{\sqrt{\bar{G}}} \cdot \sum_{\ell=1}^{q}(D \psi)_{k}^{l} \cdot \frac{\partial}{\partial \bar{x}^{\ell}} \left(det D \psi \cdot \sqrt{\bar{G}} \sum_{r=1}^{q} \sum_{s=1}^{q}\left(D \psi^{-1}\right)_{r}^{j} \bar{g}^{r s}(D \psi^{-1})_{s}^{k}\right) \right. \nonumber \\
&+\left(\sum_{l=1}^{q}(D \psi)_{j}^{l} \frac{\partial}{\partial \bar{x}^{\ell}}(D \psi)_k\right)  \left.\left(\cdot \sum_{r=1}^{q} \sum_{s=1}^{q}\left(D \psi^{-1}\right)_{r}^{j} \bar{g}^{r s}\left(D \psi^{-1}\right)_{s}^{k}\right)\right] d t \nonumber \\
=& \frac{1}{2} \sum_{l=1}^{q} \frac{1}{\sqrt{\bar{G}}} \frac{\partial}{\partial \bar{x}^{\ell}}\left(\sqrt{\bar{G}} \bar{g}^{i l}\right) d t +\frac{1}{2} \sum_{s=1}^{n} \operatorname{Tr}\left[D \psi \cdot \frac{\partial}{\partial \bar{x}^{s}}\left(D \psi^{-1}\right)\right] \cdot \bar{g}^{i s} dt \nonumber \\
&+\frac{1}{2} \sum_{j=1}^{q} \sum_{l=1}^{q} \sum_{r=1}^{q} \frac{\partial}{\partial \bar{x}^{\ell}}\left[(D \psi)_{j}^{i}\left(D \psi^{-1}\right)_{r}^j\right] \cdot \bar{g}^{rl} dt +\frac{1}{2} \sum_{l=1}^{n} \operatorname{Tr}\left[\frac{\partial}{\partial \bar{x}^{\ell}}(D \psi) \cdot\left(D \psi^{-1}\right)\right] \bar{g}^{i \ell} d t  \nonumber \\
=& \frac{1}{2} \sum_{l=1}^{q} \frac{1}{\sqrt{\bar{G}}} \frac{\partial}{\partial \bar{x}^{\ell}}\left(\sqrt{\bar{G}} \bar{g}^{i l}\right) d t 
+\frac{1}{2} \sum_{s=1}^{q} \operatorname{Tr}\left[\frac{\partial}{\partial \bar{x}^{s}}\left(D \psi \cdot D \psi^{-1}\right)\right] \cdot \bar{g}^{i s} dt +0 \nonumber \\
=&\frac{1}{2} \sum_{\ell=1}^{q} \frac{1}{\sqrt{\bar{G}}} \frac{\partial}{\partial \bar{x}^{\ell}}\left(\sqrt{\bar{G}} \bar{g}^{i \ell}\right) dt +0.
\end{align}

The $dB$ term in \eqref{apxeqn:coord} can be written in vector form as
\begin{align}
(D \psi)^{\top} \cdot g^{-\frac{1}{2}} \cdot d B.
\end{align}
Since
\begin{align*}
(D \psi)^{\top} \cdot g^{-\frac{1}{2}} \cdot d B \quad\left((D \psi)^{\top} g^{-\frac{1}{2}} d B\right)^{\top}  = \bar{g}^{-1} d t,
\end{align*}
we have
\begin{align}
(D \psi)^{\top} \cdot g^{-\frac{1}{2}} \cdot d B \sim \mathcal{N}\left(0, \bar{g}^{-1} d t\right),
\end{align}
i.e. the same distribution as $\bar{g}^{-1/2}dB$.

Conclusion: $\psi$ maps the process in $\mathbb{U} \subset \R^n$ defined by
\begin{align}
d x^{i}=\frac{1}{2} \frac{1}{\sqrt{G}} \sum_{j=1}^{q} \frac{\partial}{\partial x^{j}}\left(\sqrt{G} g^{i j}\right) d t+\left(g^{-\frac{1}{2}} d B\right)^{i}, \quad i=1, \ldots, q,
\end{align}
to the process in $\mathbb{\bar{U}} \subset \R^n$ defined by
\begin{align}
d \bar{x}^{i}=\frac{1}{2} \frac{1}{\sqrt{\bar{G}}} \sum_{j=1}^{q} \frac{\partial}{\partial \bar{x}^{j}}\left(\sqrt{\bar{G}} \bar{g}^{i j}\right) d t+\left( \bar{g}^{-\frac{1}{2}} d B\right)^{i}, \quad i=1, \ldots, q.
\end{align}
This verifies that the process defined by the above formula is coordinate independent.

As a result, with a given $dt$, simulating one step using any choice of local coordinates as above is equivalent as a step in $\mathbb{M}$.

\end{proof}

\section{GPLVM metric} \label{ax:GPLVM}
Following the description in section 3.1, the joint distribution of the $j_{th}$ dimension of the mapping $\phi$ and the $j_{th}$ column of the Jacobian can be written as  % \ref{gplvm}
\begin{align}
\begin{bmatrix}
 \phi( \X)^j\\
 \frac{ \partial \phi(x_*)^j }{\partial x} % \frac{ \partial y(x,t) }{\partial x}
\end{bmatrix}, \
\sim \mathcal N \left( 0, \begin{bmatrix}
K_{\X,\X} & \partial K_{\X,*} \\
 \partial K_{\X,*}^T & \partial^2 K_{*,*}
\end{bmatrix}  \right ).
\end{align}
We choose the covariance kernel $k$ as RBF kernel, the $(i,j)$ element of $K_{\X,\X}$ is
\begin{align*}
k(x_i,x_j) &= \gamma \exp( - \rho  || x_i - x_j ||^2 ). 
\end{align*} 
The derivatives of $K$ are:
\begin{align}
( \partial K_{\X,*} )_{i}^l &=  \frac{\partial k_{x_i,x_*}}{\partial x_*^l}  =  \rho ( x_i^{l} - x_*^{l}   )   k(x_i,x_*),   \\
(\partial^2 K_{\X,*})_{i}^{r,l} &= \frac{\partial^2 k_{x_i,x_*}}{\partial x_i^r \partial x_*^l} ,   \\
\frac{\partial^2 k_{x_i,x_*}}{\partial x_i^r \partial x_*^l}  &=  -4\rho^2  ( x_i^{r} - x_*^{r}   )  ( x_i^{l} - x_*^{l}   )   k(x_i,x_*),  & \text{if} \  \ r\neq l, \nonumber \\
\frac{\partial^2 k_{x_i,x_*}}{\partial x_i^r \partial x_*^l}   &=  2\rho ( 1 - 2\rho ( x_i^{r} - x_*^{r}   )^2  )  k(x_i,x_*),    & \text{if} \ \  r = l,  \nonumber  \\
(\partial^2 K_{*,*})^{r,l} &=   \frac{\partial^2 k_{x_*,x_*}}{\partial x_*^r \partial x_*^l}  \nonumber   \\
&= 0, &\text{if} \  \ r\neq l, \nonumber \\ 
&= 2\rho  k_{x_*,x_*} = 2\rho \gamma,   &\text{if} \ \  r = l,   \nonumber
\end{align} 
We also need the gradient of the expected metric to simulate BM trajectories. It requires computing $ \partial E[\J^T]^j / \partial x_*^l$ and $\partial \Sigma_{\J} / \partial x_*^l$.
\begin{align}
\frac{ \partial E[\J^T]^j } {\partial x_*^l} &= \frac{\partial \mu_{\J}^j}{\partial x_*^l} = \frac{ \partial  (\partial K^T_{\X,*} ) }{ \partial x_*^l }   K^{-1}_{\X,\X} \s_{:,j}, \\
\frac{\partial \Sigma_{\J} }{\partial x_*^l} &=- (\frac{\partial (\partial K_{\X,*}) }{\partial x_*^l})^T  K^{-1} \partial K_{\X,*}  - \partial K^T_{\X,*} K^{-1} \frac{ \partial ( \partial K_{\X,*} ) } { \partial x_*^l }, \\
\frac{\partial^2 k(x_i,x_*) }{\partial x_*^l\partial x_*^l} &= 2 \rho k(x_i,x_*) ( 2\rho (x_i^l - x_*^l)^2 -1 ), \nonumber \\
\frac{\partial^2 k(x_i,x_*) }{\partial x_*^l\partial x_*^r} &=  4\rho^2 (x_i^l- x_*^l)(x_i^r- x_*^r) k(x_i,x_*). \nonumber
\end{align}

\section{ Bayesian GPLVM metric} \label{ax:BGPLVM}
Following the description in section 3.2, the marginal likelihood can be derived from the the augmented joint probability as %\ref{bgplvm},
\begin{align}
%\label{apxeqn:jointprob}
p(\S ) & =  \int  \int  \int   \prod_{j=1}^p p(\s_{:}^j | \bm{\phi}_{:}^j) p( \bm{\phi}_{:}^j |  \u_{:}^j, \X) p(\u_{:}^j) p(\X) \ d \mathcal{U}     \  d\Phi \ d \X,
%p(\S) &= \int  \int  \int  p(\S, \Phi, \mathcal{U}, \X) \ d \mathcal{U}     \  d\Phi \ d \X 
\end{align} 
where $\mathcal S = \{ s_i |  i=1,\cdot \cdot \cdot, n+v \}$,  $s_i \in \R^p$, is the set of the observed data points in the original space. $\mathcal{X}= \{x_i | i= 1,\cdot \cdot \cdot, n+v \}$, $x_i \in \R^q$ is the set of latent (unobserved) variables. The mapping is denoted as $\Phi = \{ \bm{\phi}_i | i=1,\cdot\cdot\cdot, n+v \} $, $\Phi \in \R^{(n+v)\times p}$ and $ \bm{\phi}_i^j = \phi(x_i)^j$.   The inducing points are denoted by $\mathcal{U} = \{u_i| i=1,\cdot\cdot\cdot,m   \}$, $\mathcal{U} \in \R^{m\times p}$ and  $u_i = \phi(x_{ui}) \in \R^p$. The inducing points are evaluated at the pseudo-inputs $\X_u = \{ x_{ui} | i=1,\ldots,m \} $, $\X_u \in \R^{m \times q}$, in the latent space.

$p\left(\bm{\phi}^{j} \mid \u^{j}, \X, \X_u\right)$ and $p(\u^j)$  are defined as %in \eqref{apxeqn:jointprob}
\begin{align*}
p\left(\bm{\phi}^{j} \mid \u^{j}, \X, \X_u \right) & =\mathcal{N}\left(\bm{\phi}^{j} \mid K_{\X,\X_u} K_{\X_u,\X_u}^{-1} \u^{j}, K_{\X,\X}-K_{\X,\X_u} K_{\X_u,\X_u}^{-1} K_{\X_u ,X}\right), \\
p(\u^j) &= \mathcal{N} (\u^j | 0, K_{\X_u,\X_u}).
\end{align*} 
We can now apply variational inference to approximate the true posterior, $p(\Phi, \mathcal{U}, \X  | \S) = p( \Phi| \mathcal{U}, \S, \X) p( \mathcal{U}| \S, \X)p(\X|\S) $ with a variational distribution of the form
\begin{align*}
q( \Phi, \mathcal{U}, \X) = p( \Phi | \mathcal{U}, \X) q(\mathcal{U}) q(\X)= \left( \prod_{j=1}^p p(\bm{\phi}^j_{:} | \u^j_{:} , \X) q(\u_:^j)  \right)q(\X). 
\end{align*} 
The distribution $q(\X)$ is chosen to be Gaussian with variational parameters for mean and variance. Using this variational distribution and the Jensen’s inequality, we can derive the variational lower bound $\mathcal{F}$ of $\log p(\S )$ the log marginal likelihood. $p(\X)$ is chosen as Gaussian prior with identity covariance. The particular choice for the variational distribution allows us to analytically compute a lower bound.
\begin{align}
\label{apxeqn:lowbou}
\mathcal{F}( q(\X) q(\U)) &= \int q(\Phi, \U, \X) \log \frac{p ( \S, \Phi, \U, \X  ) }{ q(\Phi, \U, \X)   } d\X d\Phi \d\U  \nonumber \\
  &= \hat{ \mathcal{F} } ( q(\X), q(\U)  ) - \mathrm{KL}( q(\X) || p(\X)   ).
 \end{align} 
 Clearly, the second $\mathrm{KL}$ term in \eqref{apxeqn:lowbou} can be easily calculated since both $p(\X)$ and $q(\X)$ are Gaussian. The first term in \eqref{apxeqn:lowbou} can be written as
 \begin{align}
\hat{ \mathcal{F} } ( q(\X), q(\U)  ) & = \sum_{j=1}^p \hat{ \mathcal{F} }^j( q(\X), q(\U)   ), \nonumber \\
\hat{ \mathcal{F} }^j( q(\X), q(\U)   ) &= \int q(\u_:^j) \log \frac{    e^{\langle \log \mathcal{N}(\s_:^j |  K_{\X,\X_u} K_{\X_u,\X_u}^{-1} \u_:^{j}, \beta^{-1} I_{n+v}   )    \rangle_{q(\X)} } p(\u_:^j)  }{ q(\u_:^j)    } d\u_:^j + \mathcal{T},
 \end{align} 
 where $\mathcal{T} =  \frac{\beta}{2} \operatorname{Tr}\left(\left\langle K_{\X \X}\right\rangle_{q(X)}\right)-\frac{\beta}{2} \operatorname{Tr}\left(K_{\X_u \X_u}^{-1}\left\langle K_{\X_u \X} K_{\X \X_u}\right\rangle_{q(\X)}\right) \biggr)$. $\langle\cdot\rangle_{q(\X)}$ denotes expectation under the distribution $q(\X)$.  The expression in above equation is $\mathrm{KL}$-like quantity. And $q(\u_:^j)$ is optimally set to be proportional to the numerator inside the logarithm of the above equation which is also a Gaussian distribution. The final expression for $\hat{\mathcal{F}}$ becomes
 \begin{align}
 \label{apxeqn:vlowbou}
\begin{array}{c}
\hat{\mathcal{F}}(q(\X)) = \sum_{j=1}^p  \biggl( \log \left(\int e^{\left\langle\log \mathcal{N}\left(\s_:^{j} \mid K_{\X,\X_u} K_{\X_u,\X_u}^{-1} \u_:^{j}, \beta^{-1} I_{n+v}\right)\right\rangle_{q(X)} } p(\u_:^{j} ) d \u_:^{j}\right) \\
-\frac{\beta}{2} \operatorname{Tr}\left(\left\langle K_{\X \X}\right\rangle_{q(X)}\right)+\frac{\beta}{2} \operatorname{Tr}\left(K_{\X_u \X_u}^{-1}\left\langle K_{\X_u \X} K_{\X \X_u}\right\rangle_{q(\X)}\right) \biggr).
\end{array}
\end{align} 
This quantity can be computed in closed from since the computation of
\begin{align*} 
\psi_0 &= \operatorname{Tr}\left(\left\langle K_{\X \X}\right\rangle_{q(\X)}\right),\\
\psi_1 &=\left \langle K_{\X \X_u}\right\rangle_{q(\X)}, \\
\psi_2 &=\left \langle K_{\X_u \X} K_{\X \X_u}\right\rangle_{q(\X)},
\end{align*} 
are analytically computable for the squared exponential kernel. These quantities are referred to as $\Psi$ statistics in \cite{titsias2010}. The distribution of the inducing vairables are
\begin{align}
\label{apxeqn:vquqsig}
 q(\u_:^j) &= \mathcal{N}(\mu_{qu}^j,\Sigma_{qu}),  \\
\mu_{qu}^j&= K_{\X_u\X_u} ( \beta K_{\X_u\X_u} + \psi_2)^{-1} \psi_1^{-1} \s_:^j , \nonumber \\
\Sigma_{qu}&=\beta K_{\X_u\X_u} ( \beta K_{\X_u\X_u} + \psi_2)^{-1} K_{\X_u\X_u}. \nonumber
\end{align}
The bound can be jointly maximized over the variational parameters and the model hyperparameters by standard optimisation method such as quasi newton. The conditional probability over the Jacobian follows a Gaussian distribution.
\begin{align} 
p(  \J | \X,\S) &=  \prod^p_{j=1} \mathcal N \left( \  \partial K^T_{\X_u,*} K^{-1}_{\X_u,\X_u} \mu_{qu}^j , \  \partial^2K_{*,*} - \partial K^T_{\X_u,*} \Lambda \partial K_{\X_u,*}  \right), \nonumber \\
\Lambda &=K_{\X_u\X_u}^{-1} - K_{\X_u\X_u}^{-1} \Sigma_{qu} K_{\X_u\X_u}^{-1}, \nonumber \\
\frac{ \partial E[\J^T]^j } {\partial x_*^l} &= \frac{\partial \mu_{\J}^j}{\partial x_*^l} = \frac{ \partial  (\partial K^T_{\X_u,*} ) }{ \partial x_*^l }   K^{-1}_{\X_u,\X_u} \mu_{qu}^j, \\
\frac{\partial \Sigma_{\J} }{\partial x_*^l} &=- (\frac{\partial (\partial K_{\X_u,*}) }{\partial x_*^l})^T  \Lambda \partial K_{\X_u,*}  - \partial K^T_{\X_u,*} \Lambda \frac{ \partial ( \partial K_{\X_u,*} ) } { \partial x_*^l }.
 \end{align} 

%\section{Swiss roll kernel estimates} \label{ax:swisskernel}

%\begin{figure}[H]
% \centering
%\includegraphics[width=0.5\textwidth,height=0.5\textwidth]{metricCompare.pdf}
%    \caption{ \label{fig:metriccompare}
%       {\footnotesize 
 %{Comparison of heat kernel estimates using the analytical metric, B-GPLVM metric and GPLVM. The red solid line represents heat kernel estimates using the analytical metric. The green dashed line represents estimates using B-GPLVM metric.  }
 %    } }
%\end{figure}

%\begin{table}[H]
%\caption{\label{tb:metriccompare}Comparison of the heat kernel estimates using GPLVM metric and B-GPLVM metric. The root mean squared differences are calculated as the difference between the transition density estimates produced using the analytical metric and the GPLVM and B-GPLVM metric.}
%\vskip 0.15in
%\begin{center}
%\begin{small}
%\begin{sc}
%\begin{tabular}{lcccr}
%\hline
% Metric learning & Root mean squared differences\\
%\hline
%GPLVM & 1.2e-3\\
%B-GPLVM & 7.4e-5 \\
%\hline
%\end{tabular}
%\end{sc}
%\end{small}
%\end{center}
%\vskip -0.1in
%\end{table}

\section{Swiss roll parameterisation} \label{ax:swisspar}
The three-dimensional coordinates of the Swiss Roll can be parametrised by the radius $r$ and the width $z$. Consider the Swiss roll parametrised by
\begin{align*}
\mathbf{x}(r,z)=( r\cos r,  \ r\sin r, \  z ).
\end{align*}
To find its metric tensor, we first compute the partial derivatives
\begin{align*}
\mathbf{x}_r &= (\cos r-r\sin r, \  \sin r+r\cos r, \ 0), \ \  \mathbf{x}_z = (0,0,1).
\end{align*}
The metric tensor is given by
\begin{align*}
&\quad(\mathbf{x}_r\cdot\mathbf{x}_r)dr^2
+2(\mathbf{x}_r\cdot\mathbf{x}_z)dr\,dz
+(\mathbf{x}_z\cdot\mathbf{x}_z)dz^2 =(1+r^2)dr^2+dz^2.
\end{align*}
or in matrix form
\begin{align*}
g= \left[\begin{array}{cc}  
1+r^2 & 0 \\[0.3em]
0 & 1 \\[0.3em]
           \end{array} \right], \qquad
 g^{-1}= \left[\begin{array}{cc} 
\frac{1}{1+r^2} & 0 \\[0.3em]
0 & 1 \\[0.3em]
           \end{array} \right], \qquad
\frac{\partial g}{\partial r}= \left[\begin{array}{cc}  
2r &0 \\[0.3em]
 0 & 0 \\[0.3em]
\end{array} \right].            
\end{align*}

The determinant of the metric tensor in this case would be $1+r^2$, as $r$ grows the determinant is getting bigger. This indicates the exaggeration from the low dimensional latent space to the high dimensional original space is getting bigger. 

The BM on the Swiss Roll can be written as
\begin{align}
\label{apx:swissBM}
%dr(t) &= \frac{1}{2} \left( -g^{-1} \frac{\partial g}{\partial r} g^{-1} \right )_{11} dt + \frac{1}{4} g^{-1}_{11} tr( g^{-1} \frac{\partial g}{\partial r} )  dt + (g^{-1/2})_{11}dB_r  \\
dr(t) &= - \frac{1}{2} \frac{r}{(1+r^2)^2} dt + (1+r^2)^{-1/2} dB_r(t)  , \\
%dz(t) &=  (g^{-1/2})_{22}dB_z(t) \\
dz(t) &=  dB_z(t) . \nonumber
\end{align}

\begin{figure}[H]
 \centering
   \subfigure[ $\R^2$ GP ]{\label{swiss:r2} \includegraphics[width=0.45\textwidth,height=0.45\textwidth]{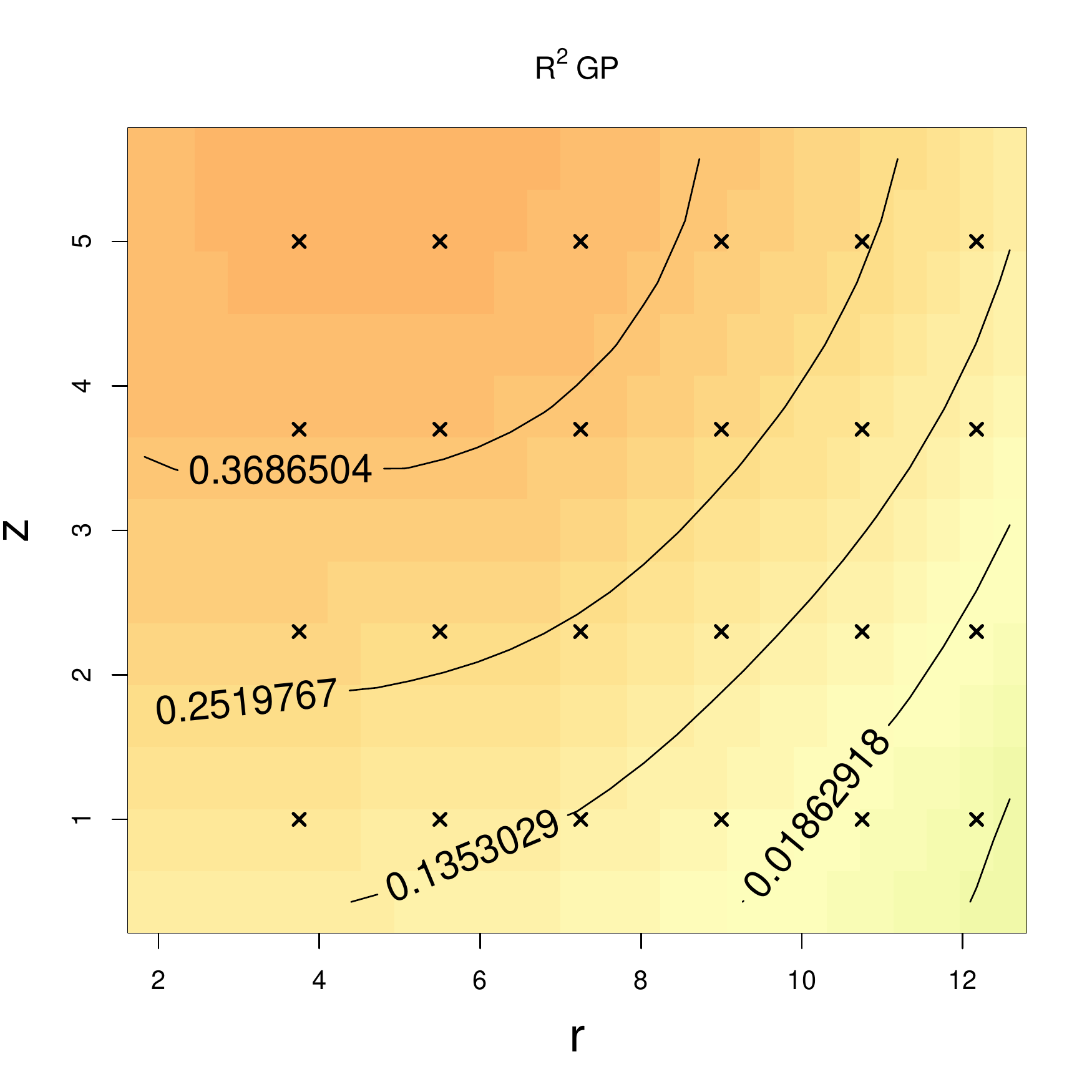}}
  \subfigure[ GL-GP ]{\label{swiss:gl} \includegraphics[width=0.45\textwidth,height=0.45\textwidth]{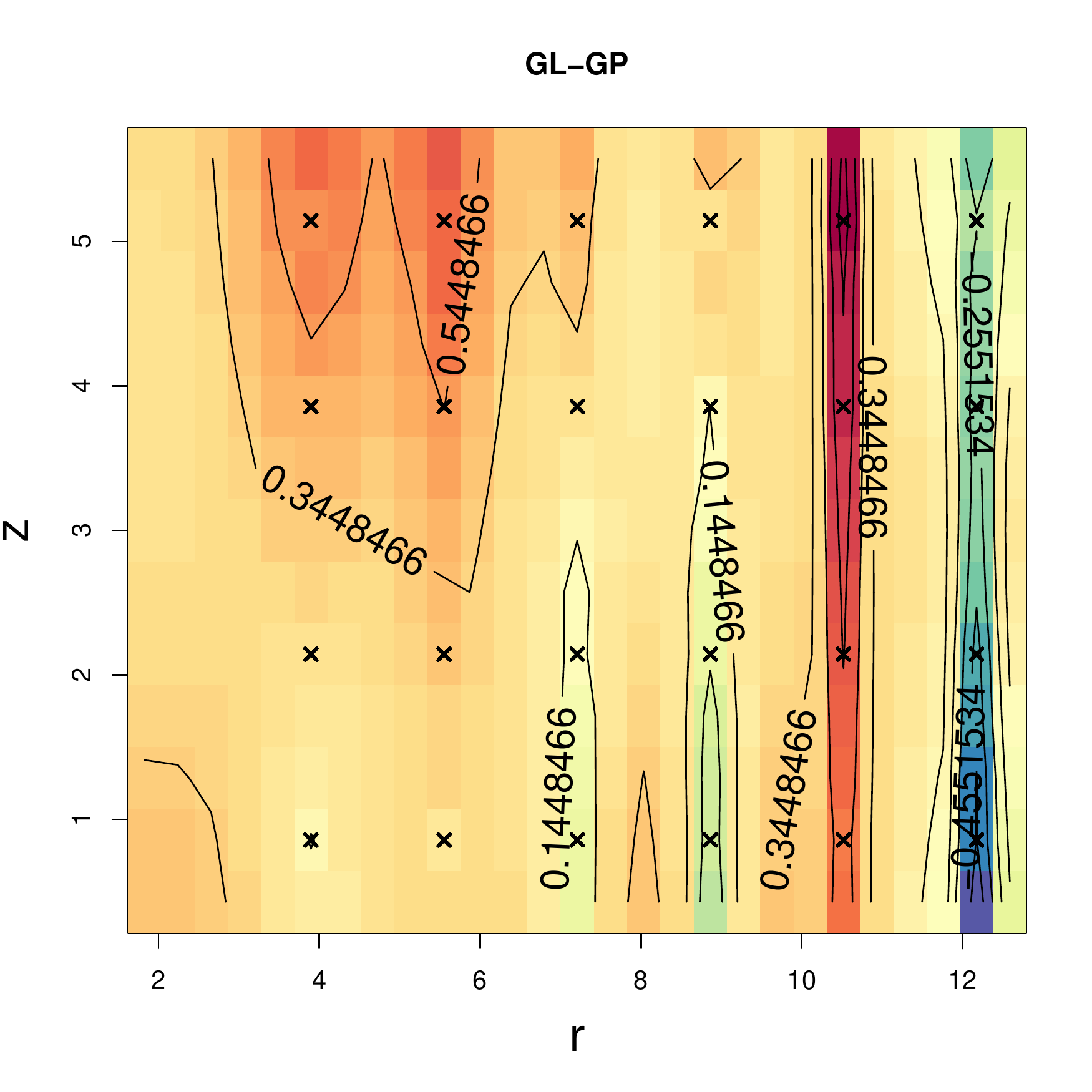}}
   \caption{ \label{swiss:appendix}
     {\footnotesize 
Prediction of $\R^2$GP and GL-GP
    } }
\end{figure}

\section{WiFi signal regression} \label{ax:wifi}

\begin{figure}[H]
    \centering
%\label{ax:wifiARD}
\includegraphics[width=0.48\textwidth,height=0.45\textwidth]{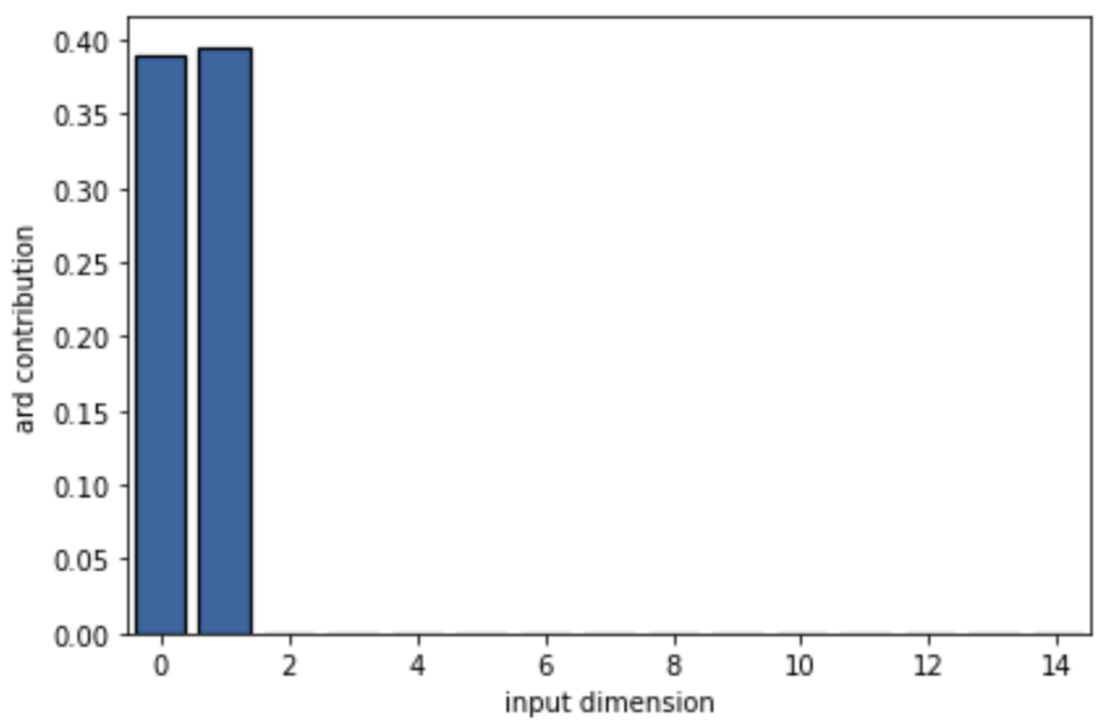}
    \caption{\label{fig:wifiARD}
   { \footnotesize 
 Automatic Relevance Determination (ARD) contributions of the WiFi signal datasets. The horizontal axis is the index of dimension. The vertical axis is the ARD contribution(or relevance) which are computed as the inverse of the squared lengthscales in B-GPLVM. If the ARD contribution is low, the corresponding dimension is less important.
    }  
 }
 %Estimating $q$, the dimension of the wifi signal manifold using the
\end{figure}

\begin{figure}[H]
    \centering
\includegraphics[width=0.45\textwidth,height=0.45\textwidth]{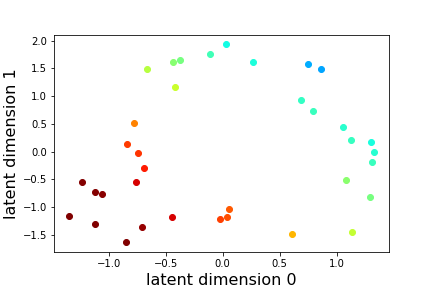}
    \caption{\label{fig:wifi2d}
   { \footnotesize
$\R^{2}$ GP prediction. The $\R^{2}$ GP is constructed using the euclidean distance of data points in the latent space of WiFi signals. It ignores the boundary and the magnification factor.
    }  
 }
\end{figure}

\section{Coil image latent space} \label{ax:coil}

\begin{figure}[H]
 \centering
\subfigure[ COIL boundary in latent space ]{%\label{coil:bound} 
\includegraphics[width=0.49\textwidth,height=0.5\textwidth]{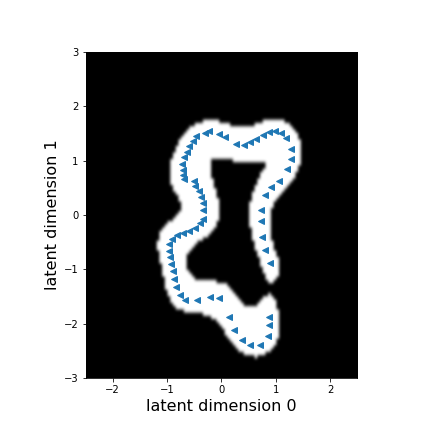}}
\subfigure[ BM trajectory ]{%\label{coil:BMpaths} 
\includegraphics[width=0.42\textwidth,height=0.46\textwidth]{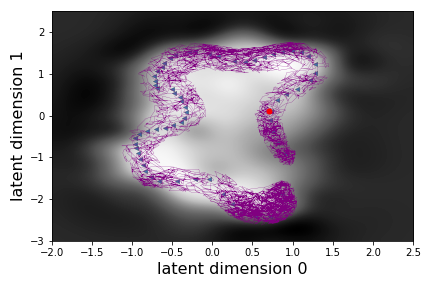}}
   \caption{ \label{coil:bound}
      {\footnotesize 
A BM sample path is simulated in the implicit manifold of COIL images. (a) The visualisation of the boundary in the latent space. The white region is within $\partial \M$. The blue triangles are the observed latent points. (b) A BM trajectory is plotted as the purple line. The red ball is the starting location. The gray background represents the magnification factor.
    } }
\end{figure}

 \begin{figure}[H]
    \centering
% \label{coil:2d} 
 \includegraphics[width=0.55\textwidth,height=0.45\textwidth]{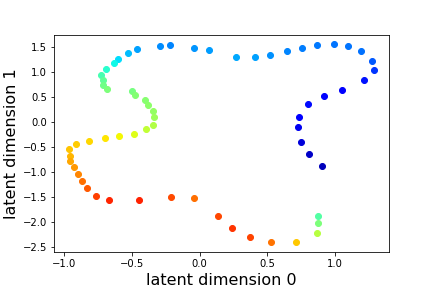}
    \caption{\label{coil:2d}
   { \footnotesize
 $\R^{2}$ GP prediction. $\R^{2}$ GP is constructed using the euclidean distance of data points in the latent space and ignoring the impact of the boundary and the magnification factor.
    }  
 }
\end{figure}

 \begin{figure}[H]
    \centering
\includegraphics[width=0.7\textwidth,height=0.6\textwidth]{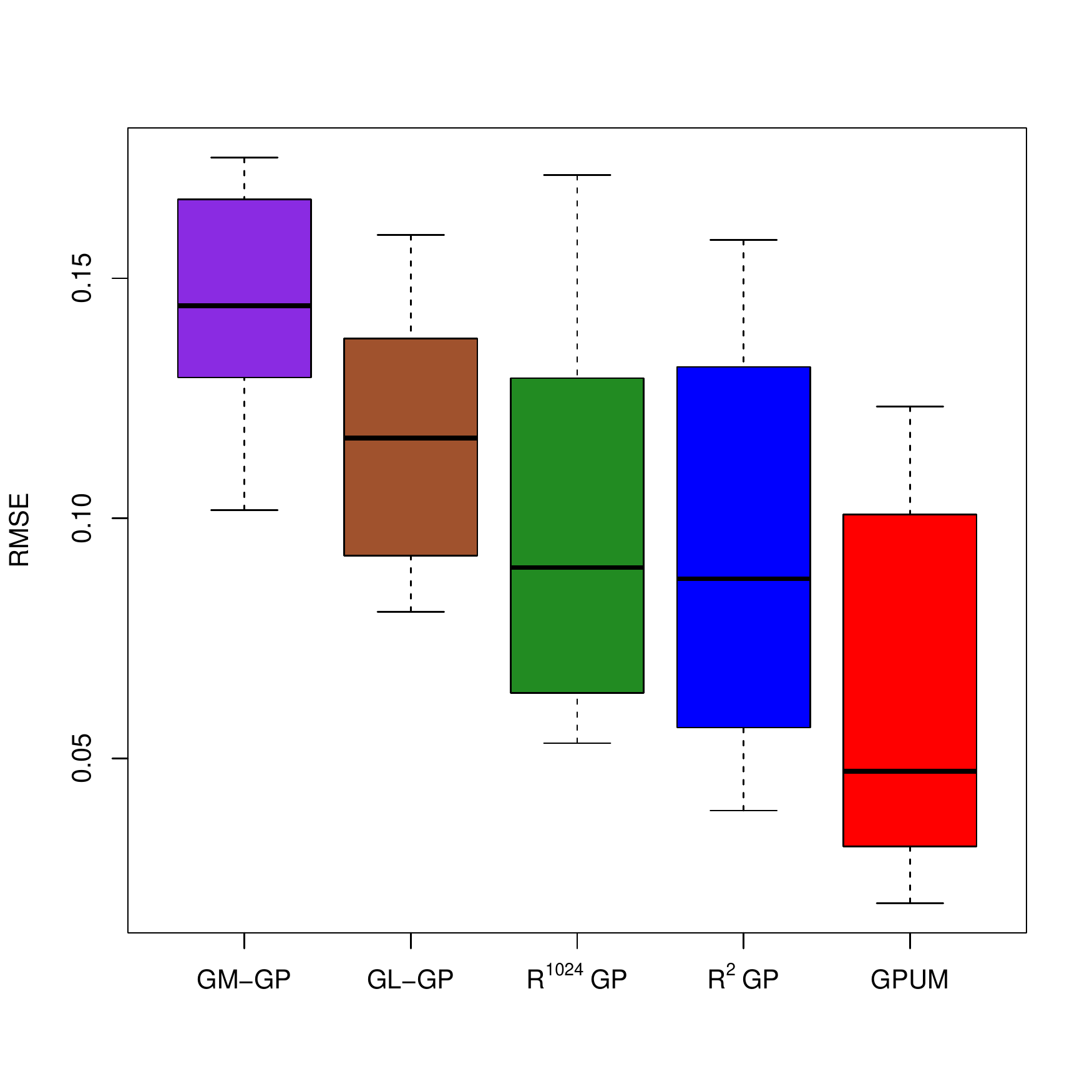}
    \caption{  \label{coil:box} 
   { \footnotesize
Boxplot of the RMSE for all methods applied in COIL images example.
    }  
 }
\end{figure}

\section{ Graph Laplacian and Graph based Gaussian Process } 
\label{ax:GL_kern}

%\subsection*{G.1 Graph Laplacian and its eigen decomposition}
Suppose $\M$ is a d-dimensional smooth closed and connected Riemannian manifold embedded in $R^p$ through $f:\M \rightarrow\R^p$. %independently, which is
Let $-\Delta$ be the Laplace-Beltrami operator of $\M$. Let $\{\lambda_i\}^{\infty}_{i=0}$ be the spectrum of $-\Delta$. We have eigenvalue: $0=\lambda_0<\lambda_1<\dots$. Denote $\phi_i$ the corresponding eigenfunction. The heat kernel has the expression: 
\[H(x,x',t)=\sum_{i=0}^{\infty}e^{-\lambda_it}\phi_i(x)\phi_i(x').\] 

Supposing we are able to recover the eigenfunctions and eigenvalues of the Laplace-Beltrami operator through the Graph Laplacian, we can recover the heat kernel via 
\[ H(x,x',t)=\sum_{i=0}^{\infty}e^{-\mu_{i,\epsilon}}\Tilde{v}_{i,\epsilon}\Tilde{v}_{i,\epsilon}^T,
\]
where $\mu$ and $\Tilde{v}$ are the eigenvalue and eigenvector the Graph Laplacian. Given a data set $X:=\{x_1,x_2,\cdots,x_{n} , \cdots,x_{n+v}\}, x_i\subset R^p$, where $n$ is the number of the labeled observation and v is the number of unlabeled grid points, we construct a kernel normalized Graph Laplacian(GL) over ${x_1,x_2,...,x_{n+v}}$ following \cite{dunson2020diffusion}'s approach. %The heat kernel of a graph can be approximated by a finite summation of the eigen pairs of the Graph Laplacian (GL).

We define a Gaussian-like kernel function: 
\[k_\varepsilon(x,x')=\exp(-\frac{||x-x'||^2_{R^p}}{4\varepsilon^2}),\]
where $\varepsilon>0$, $\varepsilon$ is the bandwidth.

The $(n+v) \times (n+v)$ affinity matrix $W$ is constructed using the normalised kernel ($\alpha-normalization$), where $q_\varepsilon(x):=\sum^{n}_{i=1}k_\varepsilon(x,x_i)$:
\[W_{ij}:=\frac{k_\varepsilon(x_i,x_j)}{q_\varepsilon(x_i)q_\varepsilon(x_j)}=\frac{k_\varepsilon(x_i,x_j)}{\sum_{i=1}^{n+v}k_\varepsilon(x,x_i)\sum_{j=1}^{n+v}k_\varepsilon(x,x_j)}.\]
%The term $q_\varepsilon$ is the kernel density evaluated at $x$. We normalized the kernel $k_\varepsilon(x,x')$ by dividing $q_\varepsilon$ to remove the impact of the non-uniform sampling density.

An $(n+v) \times (n+v)$ diagonal matrix $D$ is constructed by setting the diagonal elements as:
\[ D_{ii}=\sum_{j=1}^{n+v}W_{ij}.\]

The row stochastic transition matrix $A$ is defined as:
\[ A=D^{-1}W.\]

The Graph Laplacian (GL) matrix is: 
\[L:=\frac{A-I}{\varepsilon^2}.\]

$\Tilde{A}$ can be computed as:
\[\Tilde{A}=D^{-1/2}WD^{-1/2}.\]

$\Tilde{A}$ is diagonalizable, and the eigenvalue of $\Tilde{A}$ is the same as $A$.

%After getting the GL matrix, we discuss how to estimate the heat kernel by using it.  Suppose we have a point cloud $X:=\{x_1\dots x_{n+v}\}$, $X\in M$. 

Given the GL matrix constructed as above, denote $\mu_{i,\epsilon}$ the $i$-th eigenvalue of $-L$ with the associated eigenvector $\Tilde{v}_{i,\epsilon}$ normalized in $l^2$ norm. %$\mu_{i,n,\epsilon}$ is an approximation of the i-th eigenvalue $\lambda_i$ of $-\Delta$. 
%On a closed manifold $M$, if $K$ is fixed,

We do the following normalization of the eigenvector $\Tilde{v}_{i,n,\epsilon}$ in the $l^2$ norm. Let $N(i)=|B^{R^p}_{\epsilon}\cap(f(x_i))\{f(x_1)\dots f(x_n)\}|$ be the number of points on $\epsilon$ ball in the ambient space. We have the $l^2$ norm of $\Tilde{v}$: 
\[||\Tilde{v}||_{l^2}=\sqrt{\frac{|S^{d-1}|\epsilon^d}{d}\sum_{i=1}^{n}\frac{\Tilde{v}^2(i)}{N(i)}}.\]

We get the eigenvector after normalizing as :
\[v_{i,n,\epsilon}=\frac{\Tilde{v}_{i,n,\epsilon}}{||\Tilde{v}||_{l^2}}.\]

The heat kernel can be approximated as $GL_{kernel}$: 
\[ GL_{kernel} = \sum_{i=0}^{K-1}e^{-\mu_{i,\epsilon}}v_{i,\epsilon}v_{i,\epsilon}^T.\]
where K is the order of eigen-paires. The construction of the kernel can be summarised in Algorithm \ref{alg:GL}.

\begin{algorithm}[h]
%\setstretch{1}
\caption{GL Algorithm.}\label{alg:GL}
Algorithm inputs include $t,\epsilon,K$\\

Step (1): Construct the $(n+v)\times (n+v)$ matrix W and D as shown in Appendix \ref{ax:GL_kern} with bandwidth $\epsilon$ and points cloud $\{ x_1,.\dots,x_{n+v}\}$. We can get:
\[\Tilde{A}=D^{-1/2}WD^{-1/2}.\]

Step (2): Find the first $K-1$ eigenpairs of $\Tilde{A}$:
\[\{\alpha_{i,\epsilon},U_{i,\epsilon}\}_{i=1}^{K-1}.\]
Step (3): Suppose $\Tilde{v}_{i,\epsilon}$ is the normalized vector of $D^{-1/2}U_{i,\epsilon}$ in the $l^2$ norm, and we have: 
\[\mu_{i,\epsilon}:=\frac{1-\alpha_{i,\epsilon}}{\epsilon^2}.\]
Let $N(i)=|B^{R^p}_{\epsilon}(f(x_i))\{f(x_1)\dots f(x_n)\}|$ be the number of points on $\epsilon$ ball in the ambient space,We have the $l^2$ norm of $\Tilde{v}$: 
\[\Tilde{v}_{l^2}=\sqrt{\frac{|S^{d-1}|\epsilon^d}{d}\sum_{i=1}^{n}\frac{\Tilde{v}^2(i)}{N(i)}}.\]
For $i=1,2,\dots,K-1$, we have:
$v_{i,\epsilon}=\frac{\Tilde{v}_{i,\epsilon}}{\Tilde{v}_{l^2}}$ \\
Construct $H^K_{\epsilon,t}$ as
\[ H^K_{\epsilon,t} = \sum_{i=0}^{K-1}e^{-\mu_{i,\epsilon}}v_{i,\epsilon}v_{i,\epsilon}^T.\]
\end{algorithm}

\newpage
 
% Note: in this sample, the section number is hard-coded in. Following
% proper LaTeX conventions, it should properly be coded as a reference:

%In this appendix we prove the following theorem from
%Section~\ref{sec:textree-generalization}:

\vskip 0.2in
\bibliography{manuscript_v2}

\begin{thebibliography}{32}
\providecommand{\natexlab}[1]{#1}
\providecommand{\url}[1]{\texttt{#1}}
\expandafter\ifx\csname urlstyle\endcsname\relax
  \providecommand{\doi}[1]{doi: #1}\else
  \providecommand{\doi}{doi: \begingroup \urlstyle{rm}\Url}\fi

\bibitem[Alvarez and Lawrence(2011)]{alvarez2011}
M.~A. Alvarez and N.~D. Lawrence.
\newblock Computationally efficient convolved multiple output gaussian
  processes.
\newblock \emph{The Journal of Machine Learning Research}, 12:\penalty0
  1459--1500, 2011.

\bibitem[Anderson(1946)]{anderson1946}
T.~W. Anderson.
\newblock The non-central wishart distribution and certain problems of
  multivariate statistics.
\newblock \emph{The Annals of Mathematical Statistics}, pages 409--431, 1946.

\bibitem[Arvanitidis et~al.(2019)Arvanitidis, Hauberg, Hennig, and
  Schober]{arvanitidis2019}
G.~Arvanitidis, S.~Hauberg, P.~Hennig, and M.~Schober.
\newblock Fast and robust shortest paths on manifolds learned from data.
\newblock In \emph{The 22nd International Conference on Artificial Intelligence
  and Statistics}, pages 1506--1515. PMLR, 2019.

\bibitem[Bishop et~al.(1997)Bishop, Svens{\'e}n, and Williams]{bishop1997}
C.~M. Bishop, M.~Svens{\'e}n, and C.~K. Williams.
\newblock Magnification factors for the gtm algorithm.
\newblock 1997.

\bibitem[Bolin et~al.(2022)Bolin, Simas, and Wallin]{bolin2022}
D.~Bolin, A.~B. Simas, and J.~Wallin.
\newblock Gaussian whittle-mat$\backslash$'ern fields on metric graphs.
\newblock \emph{arXiv preprint arXiv:2205.06163}, 2022.

\bibitem[Borovitskiy et~al.(2020)Borovitskiy, Terenin, Mostowsky,
  et~al.]{borovitskiy2020}
V.~Borovitskiy, A.~Terenin, P.~Mostowsky, et~al.
\newblock Mat{\'e}rn gaussian processes on riemannian manifolds.
\newblock \emph{Advances in Neural Information Processing Systems},
  33:\penalty0 12426--12437, 2020.

\bibitem[Borovitskiy et~al.(2021)Borovitskiy, Azangulov, Terenin, Mostowsky,
  Deisenroth, and Durrande]{borovitskiy2021}
V.~Borovitskiy, I.~Azangulov, A.~Terenin, P.~Mostowsky, M.~Deisenroth, and
  N.~Durrande.
\newblock Mat{\'e}rn gaussian processes on graphs.
\newblock In \emph{International Conference on Artificial Intelligence and
  Statistics}, pages 2593--2601. PMLR, 2021.

\bibitem[Chavel(1984)]{chavel1984}
I.~Chavel.
\newblock \emph{Eigenvalues in Riemannian geometry}.
\newblock Academic press, 1984.

\bibitem[Damianou et~al.(2016)Damianou, Titsias, and Lawrence]{damianou2016}
A.~C. Damianou, M.~K. Titsias, and N.~D. Lawrence.
\newblock Variational inference for latent variables and uncertain inputs in
  gaussian processes.
\newblock \emph{The Journal of Machine Learning Research}, 17\penalty0
  (1):\penalty0 1425--1486, 2016.

\bibitem[Dunson et~al.(2020)Dunson, Wu, and Wu]{dunson2020diffusion}
D.~B. Dunson, H.-T. Wu, and N.~Wu.
\newblock Diffusion based gaussian processes on restricted domains.
\newblock \emph{arXiv preprint arXiv:2010.07242}, 2020.

\bibitem[Feragen et~al.(2015)Feragen, Lauze, and Hauberg]{feragen2015}
A.~Feragen, F.~Lauze, and S.~Hauberg.
\newblock Geodesic exponential kernels: When curvature and linearity conflict.
\newblock In \emph{Proceedings of the IEEE Conference on Computer Vision and
  Pattern Recognition}, pages 3032--3042, 2015.

\bibitem[Ferris et~al.(2007)Ferris, Fox, and Lawrence]{ferris2007}
B.~Ferris, D.~Fox, and N.~D. Lawrence.
\newblock Wifi-slam using gaussian process latent variable models.
\newblock In \emph{IJCAI}, volume~7, pages 2480--2485, 2007.

\bibitem[Hjelle and D{\ae}hlen(2006)]{hjelle2006}
{\O}.~Hjelle and M.~D{\ae}hlen.
\newblock \emph{Triangulations and applications}.
\newblock Springer Science \& Business Media, 2006.

\bibitem[Hsu(2008)]{hsu2008}
E.~P. Hsu.
\newblock A brief introduction to {B}rownian motion on a {R}iemannian manifold.
\newblock \emph{Lecture Notes}, 2008.

\bibitem[Hsu(1988)]{hsu1988}
P.~Hsu.
\newblock Brownian motion and {R}iemannian geometry.
\newblock \emph{Contemporary Mathematics}, 73:\penalty0 95--104, 1988.

\bibitem[Kingma et~al.(2019)Kingma, Welling, et~al.]{kingma2019}
D.~P. Kingma, M.~Welling, et~al.
\newblock An introduction to variational autoencoders.
\newblock \emph{Foundations and Trends{\textregistered} in Machine Learning},
  12\penalty0 (4):\penalty0 307--392, 2019.

\bibitem[Kramer(1991)]{kramer1991}
M.~A. Kramer.
\newblock Nonlinear principal component analysis using autoassociative neural
  networks.
\newblock \emph{AIChE journal}, 37\penalty0 (2):\penalty0 233--243, 1991.

\bibitem[Lawrence(2005)]{lawrence2005}
N.~Lawrence.
\newblock Probabilistic non-linear principal component analysis with gaussian
  process latent variable models.
\newblock \emph{Journal of machine learning research}, 6\penalty0
  (Nov):\penalty0 1783--1816, 2005.

\bibitem[Lawrence(2007)]{lawrence2007}
N.~D. Lawrence.
\newblock Learning for larger datasets with the gaussian process latent
  variable model.
\newblock In \emph{Artificial intelligence and statistics}, pages 243--250,
  2007.

\bibitem[{Lin} et~al.(2018){Lin}, {Mu}, {Chan}, and {Dunson}]{extrinsicGP}
L.~{Lin}, N.~{Mu}, P.~{Chan}, and D.~B. {Dunson}.
\newblock {Extrinsic {G}aussian processes for regression and classification on
  manifolds}.
\newblock \emph{Bayesian Analysis}, 2018.
\newblock In press.

\bibitem[Lin et~al.(2019)Lin, Mu, Cheung, and Dunson]{lin2019}
L.~Lin, N.~Mu, P.~Cheung, and D.~Dunson.
\newblock Extrinsic gaussian processes for regression and classification on
  manifolds.
\newblock \emph{Bayesian Analysis}, 14\penalty0 (3):\penalty0 887--906, 2019.

\bibitem[Nene et~al.(1996)Nene, Nayar, Murase, et~al.]{nene1996}
S.~A. Nene, S.~K. Nayar, H.~Murase, et~al.
\newblock Columbia object image library (coil-100).
\newblock 1996.

\bibitem[Niu et~al.(2019)Niu, Cheung, Lin, Dai, Lawrence, and Dunson]{niu2019}
M.~Niu, P.~Cheung, L.~Lin, Z.~Dai, N.~Lawrence, and D.~Dunson.
\newblock Intrinsic gaussian processes on complex constrained domains.
\newblock \emph{Journal of the Royal Statistical Society: Series B (Statistical
  Methodology)}, 81\penalty0 (3):\penalty0 603--627, 2019.

\bibitem[Qui{\~n}onero-Candela and
  Rasmussen(2005)]{QuioneroCandelaRasmussen2005}
J.~Qui{\~n}onero-Candela and C.~E. Rasmussen.
\newblock A unifying view of sparse approximate {G}aussian process regression.
\newblock \emph{Journal of Machine Learning Research}, 6, 2005.

\bibitem[Rasmussen(2004)]{Rasmussen2004}
C.~E. Rasmussen.
\newblock \emph{Gaussian Processes in Machine Learning}, pages 63--71.
\newblock Springer Berlin Heidelberg, 2004.

\bibitem[Rasmussen and Williams(2006)]{Rasmussen2006}
C.~E. Rasmussen and C.~K. Williams.
\newblock \emph{Gaussian processes for machine learning}, volume~2.
\newblock MIT press Cambridge, MA, 2006.

\bibitem[Tipping and Bishop(1998)]{tipping1998}
M.~E. Tipping and C.~M. Bishop.
\newblock Mixtures of probabilistic principal component analysers.
\newblock 1998.

\bibitem[Titsias and Lawrence(2010)]{titsias2010}
M.~Titsias and N.~D. Lawrence.
\newblock Bayesian gaussian process latent variable model.
\newblock In \emph{Proceedings of the Thirteenth International Conference on
  Artificial Intelligence and Statistics}, pages 844--851, 2010.

\bibitem[Tosi(2014)]{tosi2014}
A.~Tosi.
\newblock \emph{Visualization and interpretability in probabilistic
  dimensionality reduction models}.
\newblock PhD thesis, Universitat Polit{\`e}cnica de Catalunya, 2014.

\bibitem[Tosi et~al.(2014)Tosi, Hauberg, Vellido, and Lawrence]{tosi2014metric}
A.~Tosi, S.~Hauberg, A.~Vellido, and N.~D. Lawrence.
\newblock Metrics for probabilistic geometries.
\newblock \emph{arXiv preprint arXiv:1411.7432}, 2014.

\bibitem[Yang and Dunson(2016)]{yang2016}
Y.~Yang and D.~B. Dunson.
\newblock Bayesian manifold regression.
\newblock \emph{The Annals of Statistics}, 44\penalty0 (2):\penalty0 876--905,
  2016.

\bibitem[Zwiessele(2017)]{zwiessele2017}
M.~Zwiessele.
\newblock \emph{Bringing models to the domain: Deploying gaussian processes in
  the biological sciences}.
\newblock PhD thesis, University of Sheffield, 2017.

\end{thebibliography}

\end{document}